\PassOptionsToPackage{table}{xcolor}
\documentclass[10pt]{article} %
\usepackage{microtype}
\usepackage[preprint]{tmlr}

\usepackage{hyperref}
\usepackage{url}
\usepackage{epigraph}
\usepackage{inconsolata}
\usepackage{graphicx}
\usepackage{mathrsfs}
\usepackage{multirow}
\usepackage{booktabs}
\usepackage{enumitem}
\usepackage{soul}
\usepackage{hyperref}
\usepackage[hyperpageref]{backref}
\usepackage[font=small]{caption}
\usepackage[bottom]{footmisc}
\usepackage{pifont}

\usepackage{algorithm}
\usepackage{algpseudocode}

\definecolor{AccentColor}{HTML}{916CB4}
\definecolor{AccentColorLight}{HTML}{D8CBE4}
\definecolor{AccentColorSuperLight}{HTML}{EDE7F2}
\definecolor{CiteColor}{HTML}{7340A2}

\usepackage{subfig}
\usepackage{xkcdcolors}
\hypersetup{
    colorlinks=true,
    linkcolor=xkcdOcean,
    urlcolor=xkcdBluish,
    linktoc=all,
    citecolor=CiteColor
}%
\usepackage{wrapfig}
\usepackage{makecell}

\makeatletter
\newcommand{\LeftComment}[1]{%
  \Statex \hspace*{\ALG@thistlm}\(\triangleright\) #1}
\makeatother

\usepackage{amsmath,amsfonts,bm}

\newcommand{\dmodel}{{D_{\text{model}}}}

\def\eqref#1{Eq.~(\ref{#1})}

\def\1{\bm{1}}

\def\rvh{{\bm{h}}}
\def\rvu{{\bm{i}}}
\def\rvj{{\bm{j}}}

\def\rvm{{\bm{m}}}

\def\rvu{{\bm{u}}}
\def\rvv{{\bm{v}}}
\def\rvw{{\bm{w}}}
\def\rvx{{\bm{x}}}
\def\rvy{{\bm{y}}}

\def\vg{{\bm{g}}}

\def\vu{{\bm{u}}}

\def\mU{{\bm{U}}}
\def\mV{{\bm{V}}}
\def\mW{{\bm{W}}}
\def\mX{{\bm{X}}}
\def\mY{{\bm{Y}}}
\def\mZ{{\bm{Z}}}

\DeclareMathAlphabet{\mathsfit}{\encodingdefault}{\sfdefault}{m}{sl}
\SetMathAlphabet{\mathsfit}{bold}{\encodingdefault}{\sfdefault}{bx}{n}

\def\gQ{{\mathcal{Q}}}

\newcommand{\R}{\mathbb{R}}

\usepackage{cleveref}

\usepackage{thmtools,thm-restate}

\usepackage{xspace}
\newcommand{\ourmodel}{Chronos-2\xspace}
\newcommand{\fevbench}{\texttt{fev-bench}\xspace}
\newcommand{\gifteval}{\texttt{GIFT-Eval}\xspace}
\newcommand{\chronosbenchii}{\texttt{Chronos Benchmark II}\xspace}

\title{\textcolor{AccentColor}{\ourmodel:} From Univariate to Universal Forecasting}

\author{\name Abdul Fatir Ansari\textnormal{\textsuperscript{1}}\thanks{Equal contribution.}, Oleksandr Shchur\textnormal{\textsuperscript{1}}\footnotemark[1], Jaris Küken\textnormal{\textsuperscript{1,3}}\footnotemark[1]\,\;\thanks{Jaris Küken and Andreas Auer contributed to this work during their internships at AWS. Hao Wang and Pablo Guerron hold concurrent appointments at Amazon and their corresponding universities, and this report describes work performed at Amazon.}, Andreas Auer\textnormal{\textsuperscript{1,4}}\footnotemark[2], Boran Han\textnormal{\textsuperscript{1}},
Pedro Mercado\textnormal{\textsuperscript{1}},\\ Syama Sundar Rangapuram\textnormal{\textsuperscript{1}}, Huibin Shen\textnormal{\textsuperscript{1}}, Lorenzo Stella\textnormal{\textsuperscript{1}}, Xiyuan Zhang\textnormal{\textsuperscript{1}}, 
Mononito Goswami\textnormal{\textsuperscript{1}},\\ Shubham Kapoor\textnormal{\textsuperscript{1}}, Danielle C. Maddix\textnormal{\textsuperscript{1}}, Pablo Guerron\textnormal{\textsuperscript{2,5}}\footnotemark[2], Tony Hu\textnormal{\textsuperscript{1}}, Junming Yin\textnormal{\textsuperscript{1}}, Nick Erickson\textnormal{\textsuperscript{1}},\\
Prateek Mutalik Desai\textnormal{\textsuperscript{1}},
Hao Wang\textnormal{\textsuperscript{1,6}}\footnotemark[2], Huzefa Rangwala\textnormal{\textsuperscript{1}},
George Karypis\textnormal{\textsuperscript{1}},
\\
Yuyang Wang\textnormal{\textsuperscript{1}}\thanks{Equal advisory contribution.}, Michael Bohlke-Schneider\textnormal{\textsuperscript{1}}\footnotemark[3] \email ansarnd@amazon.de \\[10pt]
      \addr \textsuperscript{1}Amazon Web Services \textsuperscript{2}Amazon \textsuperscript{3}University of Freiburg \textsuperscript{4}Johannes Kepler University Linz \textsuperscript{5}Boston College \textsuperscript{6}Rutgers University}

\begin{document}

\maketitle

\begin{abstract}

Pretrained time series models have enabled inference-only forecasting systems that produce accurate predictions without task-specific training. However, existing approaches largely focus on univariate forecasting, limiting their applicability in real-world scenarios where multivariate data and covariates play a crucial role.
We present Chronos-2, a pretrained model capable of handling univariate, multivariate, and covariate-informed forecasting tasks in a zero-shot manner. Chronos-2 employs a group attention mechanism that facilitates in-context learning (ICL) through efficient information sharing across multiple time series within a group, which may represent sets of related series, variates of a multivariate series, or targets and covariates in a forecasting task. These general capabilities are achieved through training on synthetic datasets that impose diverse multivariate structures on univariate series.
Chronos-2 delivers state-of-the-art performance across three comprehensive benchmarks: {fev-bench}, {GIFT-Eval}, and {Chronos Benchmark II}. On {fev-bench}, which emphasizes multivariate and covariate-informed forecasting, Chronos-2’s universal ICL capabilities lead to substantial improvements over existing models. On tasks involving covariates, it consistently outperforms baselines by a wide margin. Case studies in the energy and retail domains further highlight its practical advantages. The in-context learning capabilities of Chronos-2 establish it as a general-purpose forecasting model that can be used ``as is'' in real-world forecasting pipelines.
\end{abstract}

\section{Introduction}

The advent of pretrained models (also referred to as \emph{foundation models}) has led to a paradigm shift in time series forecasting.
Instead of training a model for each time series (\emph{local models})~\citep{hyndman2018forecasting} or dataset (\emph{task-specific models})~\citep{lim2021temporal,challu2023nhits}, a single model can be trained once on large-scale time series data and then applied across different forecasting problems~\citep{ansari2024chronos,das2023decoder}.
Pretrained models greatly simplify the forecasting pipeline by eliminating the need for training from scratch for each use case. 
More remarkably, they often match or exceed the forecast accuracy of task-specific models~\citep{aksu2024gift}.

Despite these advances, a fundamental limitation persists: most pretrained models operate only on univariate data, considering solely the historical observations of a single time series to generate forecasts.
Although univariate forecasting is important, the class of real-world forecasting tasks spans far beyond it.
In practice, one may encounter tasks where multiple co-evolving time series need to be predicted simultaneously (\emph{multivariate forecasting})~\citep{banbura2010large,cohen2025time} or where forecasts depend on various external factors (\emph{covariate-informed forecasting}).
For example, cloud infrastructure metrics such as CPU usage, memory consumption, and storage I/O evolve together and benefit from joint modeling~\citep{cohen2025time}. Likewise, retail demand is heavily influenced by promotional activities, while energy consumption patterns are driven by weather conditions~\citep{petropoulos2022forecasting}.
The lack of multivariate and covariate-informed forecasting capabilities hinders the widespread adoption of pretrained models in real-world production systems. 

\begin{figure}[t!]
    \centering
    \includegraphics[width=\linewidth]{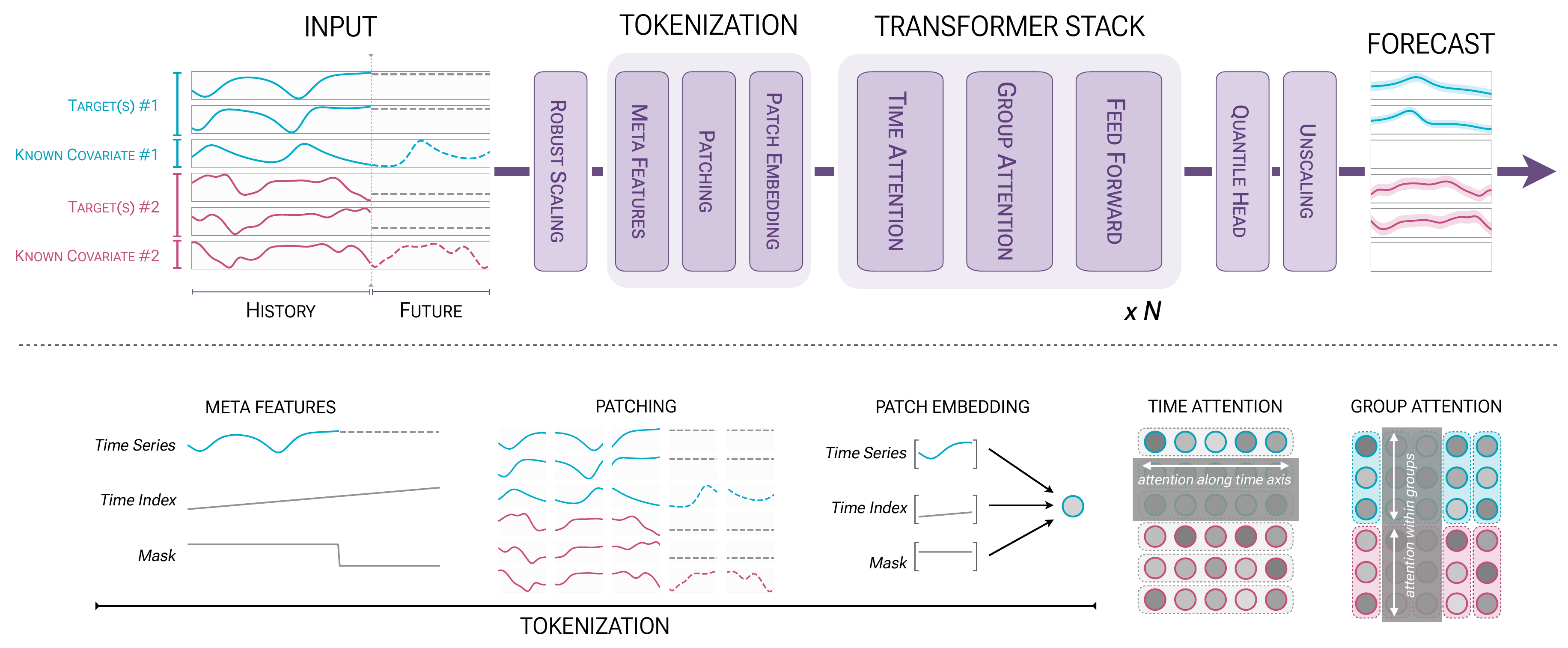}
    \caption{\textbf{\textit{The complete \ourmodel pipeline.}} Input time series (targets and covariates) are first normalized using a robust scaling scheme, after which time index and mask meta features are added. The resulting sequences are split into non-overlapping patches and mapped to high-dimensional embeddings via a residual network. The core transformer stack operates on these patch embeddings and produces multi-patch quantile outputs corresponding to the masked future patches provided as input. Each transformer block alternates between time and group attention layers: the time attention layer aggregates information across patches within a single time series, while the group attention layer aggregates information across all series within a group at each patch index. A group is a flexible notion of relatedness and may correspond to a single time series, multiple series sharing a source or metadata, variates of a multivariate series, or targets along with associated covariates. The figure illustrates two multivariate time series with one known covariate each, with corresponding groups highlighted in {\textcolor[HTML]{00ACCB}{blue}} and {\textcolor[HTML]{C64D80}{red}}. This example is for illustration only; \ourmodel supports arbitrary numbers of targets and optional covariates.}
    \label{fig:main-fig}
    \vspace{-1em}
\end{figure}

Developing \emph{universal} pretrained models that can handle both multivariate dependencies and covariates remains challenging due to two factors.
First, the heterogeneity of forecasting problems requires rethinking the model architecture.
Each downstream task differs in the number of dimensions and their semantics. Since it is impossible to know a priori how the variables will interact in an unseen task, the model must infer these interactions from the available context.
Second, high-quality pretraining data with multivariate dependencies and informative covariates is scarce.

In this work, we present \ourmodel, a pretrained model designed to handle arbitrary forecasting tasks --- univariate, multivariate, and covariate-informed --- in a \emph{zero-shot} manner.
\ourmodel leverages in-context learning (ICL) to support multivariate forecasting and arbitrary covariates, whether past-only or with known future values, real-valued or categorical.
Its enhanced ICL capabilities also improve univariate forecasting by enabling \emph{cross learning}, where the model shares information across univariate time series in the batch, leading to more accurate predictions.

At the core of \ourmodel~’s ICL capabilities is the \emph{group attention} mechanism.
It enables information exchange within groups of time series, which may represent arbitrary sets of related series, variates of a multivariate series, or targets and covariates (both past-only and known) in a forecasting task.
Rather than extending the context by concatenating targets and covariates, the group attention layer shares information within groups across the batch axis, allowing it to scale gracefully with the number of variates.
A key innovation of \ourmodel lies in our training approach:
to enable its ICL capabilities, we rely on synthetic time series data generated by imposing multivariate structure on time series sampled from base univariate generators.
The complete inference pipeline of \ourmodel including tokenization and modeling is shown in \Cref{fig:main-fig}.

Empirical evaluation on comprehensive forecasting benchmarks, including \fevbench~\citep{shchur2025fev}, \gifteval~\citep{aksu2024gift}, and \chronosbenchii~\citep{ansari2024chronos}, shows that \ourmodel achieves state-of-the-art performance.
On \fevbench, which spans a wide range of forecasting tasks --- univariate, multivariate, and covariate-informed --- \ourmodel outperforms baselines across all categories.
The largest gains are observed on covariate-informed tasks, demonstrating Chronos-2’s strength in this practically important setting.
\ourmodel offers these new capabilities while maintaining high computational efficiency, running on a single mid-range GPU (NVIDIA A10G) with a throughput of 300 time series per second.\footnote{Based on inference time for a batch of 1,024 time series with a context length of 2048 and prediction length of 64 times teps.}

The rest of the technical report is organized as follows.
\Cref{sec:background} introduces the background on time series forecasting and existing forecasting methods with a special focus on pretrained models.
In \Cref{sec:method}, we describe the architecture of \ourmodel and discuss its training and inference pipelines.
\Cref{sec:data} briefly discusses the training corpus of \ourmodel.
In \Cref{sec:experiments}, we present our main results on three forecasting benchmarks, case studies on energy and retail domains, and ablations.
We conclude the report and discuss potential future work in \Cref{sec:discussion}.

\section{Background and Related Work}
\label{sec:background}

\newcommand{\yass}{\textcolor[HTML]{916CB4}{\ding{51}}}
\newcommand{\bruh}{\textcolor[HTML]{B44C43}{\ding{55}}}
\begin{table}[tb]
    \centering
    \resizebox{\textwidth}{!}{%
    \begin{tabular}{lccccccc}
        \toprule
        \textbf{Model} & \begin{tabular}[c]{@{}c@{}}\textbf{Univariate}\\\textbf{Forecasting}\end{tabular} & \begin{tabular}[c]{@{}c@{}}\textbf{Multivariate}\\\textbf{Forecasting}\end{tabular} & \begin{tabular}[c]{@{}c@{}}\textbf{Past-Only}\\\textbf{Covariates}\end{tabular} & \begin{tabular}[c]{@{}c@{}}\textbf{Known}\\\textbf{Covariates}\end{tabular} & \begin{tabular}[c]{@{}c@{}}\textbf{Categorical}\\\textbf{Covariates}\end{tabular} & \begin{tabular}[c]{@{}c@{}}\textbf{Cross}\\\textbf{Learning}\end{tabular} & \begin{tabular}[c]{@{}c@{}}\textbf{Memory}\\\textbf{Scaling}\end{tabular} \\
        \midrule
        \textbf{Chronos-2} & \yass & \yass & \yass & \yass & \yass & \yass & $\mathcal{O}(V)$ \\
        Toto-1.0 & \yass & \yass & \yass & \bruh & \bruh & \bruh & $\mathcal{O}(V)$ \\
        TabPFN-TS & \yass & \bruh & \bruh & \yass & \yass & \bruh & $\mathcal{O}(V)$ \\
        COSMIC & \yass & \bruh & \yass & \yass & \bruh & \bruh & $\mathcal{O}(V^2)$ \\
        Moirai-1.0 & \yass & \yass & \yass & \yass & \bruh & \bruh & $\mathcal{O}(V^2)$ \\
        \cmidrule(lr){1-8}
        Chronos-Bolt & \yass & \bruh & \bruh & \bruh & \bruh & \bruh & - \\
        Moirai-2.0 & \yass & \bruh & \bruh & \bruh & \bruh & \bruh & - \\
        Sundial & \yass & \bruh & \bruh & \bruh & \bruh & \bruh & - \\
        TimesFM-2.5 & \yass & \bruh & \bruh & \bruh & \bruh & \bruh & - \\
        TiRex & \yass & \bruh & \bruh & \bruh & \bruh & \bruh & - \\
        \bottomrule
    \end{tabular}%
    }
    \caption{Comparison of capabilities of pretrained forecasting models. \textit{Past-Only Covariates}: support for covariates only observed in the past;
    \textit{Known Covariates}: support for covariates whose future values are known;
    \textit{Categorical Covariates}: support for nominal features in the covariates;
    \textit{Cross Learning}: support for in-context learning across related time series;
    \textit{Memory Scaling}: inference memory requirements with respect to the total number of variates $V$ (including both targets and covariates).}
    \label{tab:model_comparison}
\end{table}

Time series forecasting aims to predict future values of a temporal sequence given historical observations.
Formally, let $\mY_{1:T} = \left[\rvy_1, \dots, \rvy_{T}\right]$ denote a historical time series of length $T$, where each observation $\rvy_t \in \mathbb{R}^D$ can either be univariate ($D=1$) or multivariate ($D>1$).
Given this historical context, the goal is to predict the next $H$ time steps $\mY_{T+1:T+H}$, where $H$ defines the forecast horizon.
Forecasts may be supported by covariates (also known as \emph{exogenous variables}) $\mX_{1:T+H} = \left[\rvx_1, \dots, \rvx_{T+H}\right]$, where $\rvx_t \in \mathbb{R}^M$ represents additional information that can span both historical ($t \leq T$) and future ($t > T$) time steps.
The task itself can be defined as either \emph{point forecasting}, where the objective is to predict a single future value at each time step, or \emph{probabilistic forecasting}, where the objective is to estimate the conditional distribution $\mathcal{P}(\mY_{T+1:T+H} \mid \mY_{1:T}, \mX_{1:T+H})$ in order to capture forecast uncertainty.
\emph{Zero-shot forecasting} refers to the setting in which a model generates forecasts for a previously unseen time series datasets without requiring any
additional training, adaptation, or fine-tuning.

Forecasting methods preceding the pretrained model paradigm can be broadly divided into local and global models.
Local models fit one set of parameters for each time series in the dataset. These include classical approaches such as ARIMA, Exponential Smoothing \citep{hyndman2018forecasting}, and Theta \citep{assimakopoulos2000theta}.
In contrast, global models share their parameters across all time series within a specific dataset. Deep learning approaches in this category have become increasingly common over the last decade.
Notable examples of global models include recurrent neural networks (RNN) like DeepState \citep{rangapuram2018deep}, DeepAR \citep{salinas2020deepar}, and TimeGrad \citep{rasul2021AutoregressiveDD}; stacked architectures such as N-BEATS \citep{Oreshkin2020N-BEATS} and N-HITS \citep{challu2023nhits}; and transformer-based architectures like TFT \citep{lim2021temporal} and PatchTST \citep{Nie2023PatchTST}.

Pretrained forecasting models have recently emerged as a new paradigm in time series forecasting.
While earlier work already demonstrated limited transfer learning capabilities for forecasting \citep{Orozco2020, Oreshkin2021, jin2022domain, Nie2023PatchTST}, pretrained models adopt principles similar to large language models (LLMs) and enable zero-shot generalization on diverse datasets.
Initial attempts focused on directly adapting language models to time series tasks \citep{gruver2023LLMTime,jin2024timellm}, whereas more recent approaches primarily borrow architectural concepts from LLMs but pretrain them on time series data~\citep{das2023decoder, garza2024timegpt1, ansari2024chronos}.

The majority of pretrained models are limited to univariate forecasting \citep{rasul2023lagllama, das2023decoder, ansari2024chronos, liu2025sundial, auer2025tirex}, treating each dimension independently in multivariate scenarios and ignoring covariates.
Notable exceptions include Moirai-1 \citep{woo2024unified} and Toto \citep{cohen2025time}, which incorporate multivariate structure into their architectures. Moirai-1 supports multivariate inputs but flattens them internally, which limits scalability to high-dimensional cases. Toto introduces a cross-variate attention mechanism but does not support known or categorical covariates.
COSMIC \citep{auer2025zero} advances covariate utilization through synthetic augmentations but remains restricted to univariate targets. TabPFN-TS \citep{hoo2025tables}, a tabular foundation model adapted for time series, can incorporate known covariates but it does not model past-only covariates or multivariate targets.
Despite these advances, empirical analyses show that most approaches provide only marginal benefits over univariate models \citep{zukowska2024towards, auer2025zero}, indicating that jointly modeling multiple variates and integrating covariates effectively in a zero-shot setting remains an open challenge.

Our approach addresses this gap with a \emph{group attention} mechanism, which generalizes ideas from cross-attention architectures for multivariate forecasting \citep{zhang2023crossformer, rao2021msa, arnab2021vivit} and cross-learning across multiple univariate series~\citep{das2024context}.
Unlike prior approaches, group attention operates over groups of related time series and naturally accommodates diverse forecasting setups, including univariate, multivariate, and covariate-informed tasks, within a unified framework without requiring architectural changes or task-specific adaptations.
\Cref{tab:model_comparison} compares the capabilities of \ourmodel with those of existing pretrained models.

\section{The \ourmodel Model}
\label{sec:method}

In this section, we introduce the \ourmodel model. We begin with scaling and tokenization, followed by the model's architecture including the group attention mechanism which enables \ourmodel's in-context learning capabilities.
Subsequently, we discuss the training and inference pipelines of \ourmodel. The complete inference pipeline of \ourmodel is visualized in Figure \ref{fig:main-fig}.

\subsection{Scaling and Tokenization}

\paragraph{Input Construction.}
The model operates on two inputs derived from the target $\mY_{1:T}$ and covariates $\mX_{1:T+H}$.
We concatenate all historical values into $\mV = [\rvv_1, \dots, \rvv_T]$, where each $\rvv_t \in \mathbb{R}^{D+M}$ consists of the target observation $\rvy_t$ and the corresponding covariate vector $\rvx_t$.
Similarly, we define the future values as $\mW = [\rvw_{T+1}, \dots, \rvw_{T+H}]$, where $\rvw_t \in \mathbb{R}^{D + M}$ contains known future covariate values $\rvx_{t}$ when available, while the entries corresponding to targets and past-only covariates are set to missing values.

Categorical covariates in $\mX_{1:T+H}$ are transformed into real-valued representations before being concatenated into $\mV$ and $\mW$. 
For univariate targets, we apply target encoding  \citep{pedregosa2011scikit,micci2001preprocessing}, which maps each category to a numerical value based on its relationship with the target. 
For multivariate targets, the model falls back to ordinal encoding, assigning a unique integer to each category.

\paragraph{Robust Scaling.} The input values, $\mV$ and $\mW$, may be at an arbitrary scale, so our tokenization pipeline begins by normalizing the series.
We adopt \emph{standardization}, a widely used normalization method in the literature, and introduce an additional step: applying the $\sinh^{-1}$ transformation to the standardized values.
This log-like transformation further stabilizes variance and reduces the influence of outliers on the objective function.
It has been used in econometrics~\citep{burbidge1988alternative} and energy price forecasting~\citep{uniejewski2018efficient} literature for handling extreme values.
Formally, each historical value $v_{t,d}$ and the future value $w_{t,d}$ are normalized as
\begin{align}
\tilde{v}_{t,d} &= \sinh^{-1}\!\left(\frac{v_{t,d} - \mu_d}{\sigma_d}\right) &\quad& \text{for } t \in \{1,\dots,T\}, \label{eq:normalization}\\
\tilde{w}_{t,d} &= \sinh^{-1}\!\left(\frac{w_{t,d} - \mu_d}{\sigma_d}\right) &\quad& \text{for } t \in \{T+1,\dots,T+H\},
\end{align}

where $\mu_d$ and $\sigma_d$ are the mean and standard deviation of the historical values $[v_{1,d}, ..., v_{T,d}]$, respectively.
Any missing values in $\mV$ are excluded when computing $\mu_d$ and $\sigma_d$.
The normalized historical values $\tilde{\mV}$ and future values $\tilde{\mW}$ are concatenated to construct the input matrix $\mU = [\tilde{\mV}, \tilde{\mW}] \in \R^{(T+H) \times (D +M)}$.

\paragraph{Meta Features.} 
During tokenization, each dimension of $\mU$ is processed independently by the model. 
To describe the tokenization procedure, consider a single column $\vu_d = [u_{1,d}, \dots, u_{T+H,d}]^\top$ corresponding to one target or covariate dimension $d$.
Two additional meta features are appended to each column: a time index and a mask. The time index
$
    \rvj = \left[-\frac{T}{C}, -\frac{T-1}{C}, \dots, 0, \dots, \frac{H-1}{C}\right]
$
encodes the relative position of each time step,
where $C$ is the maximum context length supported by the model.
It provides explicit information about temporal ordering to the model which is beneficial when using patch-based inputs.
The mask $\rvm_d$ is a binary indicator equal to 1 when the value is observed, and 0 otherwise.
It serves two purposes: indicating which values are missing in the historical context and specifying which input dimensions correspond to future-known covariates.
After construction of the mask, all missing values in $\vu_d$ are replaced with zeros.

\paragraph{Patching and Embedding.} The input $\rvu_d$ with the corresponding meta features, $\rvj$ and $\rvm_d$, are split into non-overlapping patches of length $P$~\citep{Nie2023PatchTST}.
The context and future sections of the time series and meta features are split into patches separately.
When $T$ and $H$ are not multiples of $P$, zero padding is applied on the left (context) or right (future).
Let $\overline{\rvu}_p$, $\overline{\rvj}_p$, and $\overline{\rvm}_p$ denote the $p$-th patches of the input, time index, and mask, respectively.
These are concatenated and mapped into the embedding space using a residual network, $f^{\mathrm{in}}_\phi: \R^{3P} \to \R^{\dmodel}$,
\begin{equation}
    \rvh_{p} = f^{\mathrm{in}}_\phi\left(\left[\overline{\rvu}_p, \overline{\rvj}_p, \overline{\rvm}_p\right]\right),
\end{equation}
where $\phi$ denotes parameters of the residual network and $\dmodel$ is the hidden dimension of the transformer model.
Between the patch embeddings of the context and future, we include a special \texttt{REG} token which serves both as a separator token and an attention sink~\citep{xiao2023efficient}. 

\subsection{Architecture}
\ourmodel is an encoder-only transformer~\citep{vaswani2017attention} model which closely follows the design of the T5 encoder~\citep{raffel2020exploring}.
In the following, we discuss the key architectural components of \ourmodel.

\paragraph{Time Attention.} The time attention layer is the usual attention layer found in typical sequence models. It applies self-attention along the temporal axis and aggregates information across patches of the same input dimension. We replace relative position embeddings used in the self-attention layers of the original T5 model with rotary position embeddings (RoPE)~\citep{su2024roformer} which have become the de-facto standard for position embeddings in modern transformer-based models~\citep{touvron2023llama}.

\paragraph{Group Attention.}
We introduce a \emph{group attention} layer into the transformer stack, which is central to enabling the in-context learning capabilities of \ourmodel. This layer aggregates information across time series that belong to the same group at a given patch index. A group refers to a set of related time series and may refer to different things depending on the forecasting task. For example, a group may consist of:
\begin{itemize}[noitemsep,nosep]
    \item \emph{a single time series}: the minimal grouping where the model makes univariate predictions without referring to other time series in the batch.
    \item \emph{a set of time series with shared source or metadata}: this grouping enables the model to perform cross learning across items by making joint predictions for related time series (also referred to as \emph{few-shot learning}) instead of generating univariate forecasts by solely taking the histories of individual time series into account. Sharing information between related time series could be especially helpful when all or some (cold start scenario) time series have short histories and when the characteristics of the downstream dataset differ considerably from the training data distribution.
    \item \emph{a set of variates with shared dynamics}: this grouping enables multivariate forecasting where the model jointly predicts all variates with shared dynamics.
    \item \emph{a set of target(s), past-only covariates and known covariates}: the most general case where the model forecasts targets while taking covariates into account.  
\end{itemize}

Within a batch of size $B$, multiple groups of varying sizes are possible, each identified by group IDs $\vg$, a vector of length $B$.
Internally, the group attention layer maps these IDs to a two-dimensional attention mask, ensuring that aggregation occurs only within groups and not across them. Since time series within a group lack a natural ordering, the group attention layer omits positional embeddings.

\paragraph{Quantile Head.}

After a sequence of alternating time and group attention layers, the embeddings of future patches of the $D$ target dimensions are passed through a residual block to produce the direct multi-step quantile forecast $\hat{\mZ} \in \mathbb{R}^{H \times D \times |\gQ|}$.
By producing forecasts for multiple target patches within a single forward pass, the model can efficiently generate predictions over long forecast horizons.
\ourmodel predicts a set of 21 quantiles $\gQ = \{0.01, 0.05, 0.1, \dots, 0.9, 0.95, 0.99\}$. 
This results in a richer representation of the predictive distribution compared to the 9-quantile grid $\{0.1, 0.2, ..., 0.9\}$ commonly used in existing pretrained models.
The inclusion of extreme quantiles ($0.01$ and $0.99$) improves coverage of rare events and enhances the model’s applicability to tasks such as anomaly detection and risk-aware forecasting.

\subsection{Training}
\label{sec:training}

During training, batches are constructed to include heterogeneous forecasting tasks: univariate forecasting, multivariate forecasting (which also covers tasks with past-only covariates), and multivariate forecasting with known covariates. Each task is characterized by the number of target dimensions $D$, the number of covariates $M$, and the role of each dimension (target, past-only covariate, or known covariate).
A unique group ID is assigned to each task, and the combination of group IDs $\vg$ with whether the future input $\mW$ is observed allows the model to infer the specific forecasting setup.

The model is trained using the quantile regression objective
\begin{equation}
   \sum_{q \in \gQ} \Big(q \cdot \max(z - \hat{z}^q, 0) + (1 - q) \cdot \max(\hat{z}^q - z, 0)\Big),
\end{equation}
where $\hat{z}^q$ is the forecast at quantile level $q$, and $z$ is the corresponding target value normalized as in~\eqref{eq:normalization}. 
The loss is averaged over all forecast steps and items in the batch and is computed only on target dimensions, with entries corresponding to known covariates or missing target values excluded from the objective.
The number of output patches is randomly sampled for each batch during training. 

Training proceeds in two stages. First, the model is pretrained with a maximum context length of 2048 and a low number of maximum output patches. In the second stage, the context length is extended to 8192, and the maximum number of sampled output patches is increased. Longer contexts enable the model to capture long-term seasonalities in high-frequency time series, while multi-patch outputs allow for long-horizon forecasts without relying on heuristics.

\subsection{Inference}
\label{sec:inference}

\begin{table}[h]
    \centering
\begin{tabular}{p{7.5cm}cc}
\toprule
\textbf{Task Type} & \textbf{Group IDs} $\vg$ & \textbf{Future Inputs} $\mW$ \\
\midrule
Univariate Forecasting \par (\emph{3 independent series}) & 
$\vg = (1, 2, 3)$ & 
$\mW = \begin{bmatrix}
\ast & \dots & \ast \\
\ast & \dots & \ast \\
\ast & \dots & \ast
\end{bmatrix} \in \mathbb{R}^{3 \times H}$ \\
\cmidrule{1-3}
Multivariate Forecasting \par (\emph{3 targets}) & 
$\vg = (1, 1, 1)$ & 
$\mW = \begin{bmatrix}
\ast & \dots & \ast \\
\ast & \dots & \ast \\
\ast & \dots & \ast
\end{bmatrix} \in \mathbb{R}^{3 \times H}$ \\
\cmidrule{1-3}
Forecasting with Covariates \par (\emph{1 target, 1 past-only covariate, 2 known covariates}) & 
$\vg = (1, 1, 1, 1)$ & 
$
\mW = \begin{bmatrix}
\ast & \dots & \ast \\
\ast & \dots & \ast \\
x_{T+1,3} & \dots & x_{T+H,3} \\
x_{T+1,4} & \dots & x_{T+H,4}
\end{bmatrix} \in \mathbb{R}^{4 \times H}
$ \\
\bottomrule
\end{tabular}
    \caption{Diverse forecasting tasks can be solved by specifying group IDs and future inputs appropriately. Here, $\vg$ and $\mW$ denote the group IDs and future values provided to the model. Future inputs for targets and past-only covariates are masked as missing values, denoted as $\ast$. The examples use fixed numbers of variates for clarity, but \ourmodel can handle arbitrary dimensions.}
    \label{tab:forecasting-tasks}
\end{table}

Forecasts are generated by de-normalizing the model predictions $\hat{z}_{t,d}^q$ and inverting \eqref{eq:normalization}.
Formally, the quantile head output $\hat{z}_{t,d}^q$ is transformed as
\begin{align}
    \hat{y}_{t,d}^q &= \mu_d + \sigma_d \cdot \sinh({\hat{z}_{t,d}^q}),
\end{align}
to obtain the prediction $\hat{y}_{t,d}^q$ of the quantile level $q$ at time step $t$ along the target dimension $d$.

During inference, multiple time series in a batch can be grouped to solve different forecasting tasks:
\begin{itemize}[nosep,noitemsep]
    \item \emph{univariate forecasting}: each item in the batch is assigned a unique group ID. This ensures that the model makes independent predictions for each time series in the batch.
    \item \emph{multivariate forecasting}: each variate which belongs to the same multivariate series is assigned the same group ID with variates from different multivariate series having distinct group IDs. This allows the model to share dynamics information between different variates of a multivariate time series.
    \item \emph{forecasting with covariates}: all target(s), past-only and known covariates belonging to the same task are assigned the same group ID. The future inputs $\mW$ corresponding to known covariates contain their known future values. The predictions generated by the model for covariates are ignored.
\end{itemize}

\Cref{tab:forecasting-tasks} summarizes how group IDs and future inputs must be specified to solve different forecasting tasks. In addition to these, \ourmodel can also be used in the \emph{full cross learning} mode where each item in the batch is assigned the same group ID regardless of whether the item is a target, a past-only covariate or a known covariate.
Since each item belongs to the same group, the model shares information across items in the batch and makes joint predictions for the entire batch.

\section{Training Data}
\label{sec:data}

For a generalist pretrained model such as \ourmodel, the training data often plays a more decisive role than the model’s specific architecture. Although recent efforts have expanded the availability of large-scale time series datasets~\citep{woo2024unified,ansari2024chronos,aksu2024gift}, they primarily contain univariate data. To overcome this limitation and endow \ourmodel with in-context learning capabilities, we relied extensively on synthetic data.

\subsection{Univariate Data}

We incorporated select datasets from the Chronos~\citep{ansari2024chronos} and GIFT-Eval~\citep{aksu2024gift} pretraining corpora into \ourmodel's training corpus. The full list of datasets is provided in Table \ref{tab:real-uni-datasets} (Appendix). To further enhance data diversity, we generated synthetic data using two approaches:

\begin{itemize}[nosep,noitemsep]
\item \textbf{TSI (Trend, Seasonality, and Irregularity)}: based on \citet{bahrpeyma2021methodology}, this generator produces diverse synthetic series by randomly constructing and combining different trend, seasonality, and irregularity components.
\item \textbf{TCM (Temporal Causal Model)}: this generator samples random causal graphs from a temporal causal model \citep{runge2023causal}, from which time series are generated via autoregression.
\end{itemize}

\subsection{Multivariate Data}

For multivariate and covariate-informed tasks, we relied entirely on synthetic data. To enable a broad class of multivariate structures, we introduce the concept of \emph{multivariatizers}. A multivariatizer samples multiple time series from base univariate generators and imposes dependencies among them to create multivariate dynamics. As base univariate generators, we employed a diverse set including autoregressive (AR) models, exponential smoothing (ETS) models, TSI, and KernelSynth~\citep{ansari2024chronos}.

We used two broad classes of multivariatizers:

\begin{itemize}[nosep,noitemsep]
\item \emph{Cotemporaneous multivariatizers} apply linear or nonlinear transformations at the same time step across time series sampled from the base univariate generators. This introduces instantaneous correlations between the time series resulting in a multivariate time series.
\item \emph{Sequential multivariatizers} induce dependencies across time, generating richer multivariate properties such as lead–lag effects and cointegration.
\end{itemize}

The multivariate time series generated from the multivariatizers were used to construct both multivariate tasks (where all variates must be predicted) and covariate-informed tasks, where a subset of variates was randomly designated as known covariates.

\section{Experiments}
\label{sec:experiments}

In this section, we present empirical results, beginning with an evaluation of \ourmodel against state-of-the-art approaches across three comprehensive benchmarks (\Cref{sec:bench-results}). We then demonstrate the gains achieved through in-context learning on univariate, multivariate, and covariate-informed forecasting tasks (\Cref{sec:icl-improvements}). Next, we examine \ourmodel's performance on tasks from the energy and retail domains, where covariates are often important for accurate forecasting (\Cref{sec:domain}). Finally, we report results for ablated variants of \ourmodel (\Cref{sec:ablations}), including a smaller model, a version trained only on synthetic data, and the model prior to long-context post-training.

\subsection{Benchmark Results}
\label{sec:bench-results}

\begin{table}[h]
    \centering
    \resizebox{\textwidth}{!}{
    \begin{tabular}{lrrrrr}
\toprule
\textbf{Model} & \textbf{Avg. Win Rate (\%)} & \textbf{Skill Score (\%)} & \textbf{Median runtime (s)} & \textbf{Leakage (\%)} & \textbf{\#Failures} \\
\midrule
\rowcolor{AccentColorLight} Chronos-2 & \bfseries 90.7 & \bfseries 47.3 & 3.6 & 0 & 0 \\
TiRex & 80.8 & 42.6 & 1.4 & 1 & 0 \\
TimesFM-2.5 & 75.9 & 42.3 & 16.9 & 8 & 0 \\
Toto-1.0 & 66.6 & 40.7 & 90.7 & 8 & 0 \\
COSMIC & 65.6 & 39.0 & 34.4 & 0 & 0 \\
Moirai-2.0 & 61.1 & 39.3 & 2.5 & 28 & 0 \\
\rowcolor{AccentColorSuperLight} Chronos-Bolt & 60.3 & 38.9 & 1.0 & 0 & 0 \\
TabPFN-TS & 59.3 & 39.6 & 305.5 & 0 & 2 \\
Sundial & 41.0 & 33.4 & 35.6 & 1 & 0 \\
Stat. Ensemble & 40.4 & 20.2 & 690.6 & 0 & 11 \\
AutoARIMA & 35.2 & 20.6 & 186.8 & 0 & 10 \\
AutoETS & 29.1 & -26.8 & 17.0 & 0 & 3 \\
AutoTheta & 21.8 & 5.5 & 9.3 & 0 & 0 \\
SeasonalNaive & 14.5 & 0.0 & 2.3 & 0 & 0 \\
Naive & 7.8 & -45.4 & 2.2 & 0 & 0 \\
\bottomrule
\end{tabular}

    }
    \caption{\textbf{\textit{\fevbench results.}} The average win rate and skill score are computed with respect to the scaled quantile loss (SQL) metric. Higher values are better for both. \ourmodel outperforms all existing pretrained models by a substantial margin on this benchmark that includes univariate, multivariate, and covariate-informed forecasting tasks. Baseline results and the imputation strategy for handling data leakage in certain tasks are both taken from \citet{shchur2025fev}. Results for additional forecasting metrics are provided in \Cref{tab:fev-results-mase,tab:fev-results-wql,tab:fev-results-wape} (Appendix).}
    \vspace{-1em}
    \label{tab:fev-results-sql}
\end{table}

We evaluated the \emph{base} \ourmodel model with 120M parameters on three comprehensive forecasting benchmarks: \fevbench~\citep{shchur2025fev}, \gifteval~\citep{aksu2024gift}, and \chronosbenchii~\citep{ansari2024chronos}.
To contextualize its performance, we compared it against state-of-the-art time series foundation models that achieved the strongest results on these benchmarks.
These include TiRex~\citep{auer2025tirex}, TimesFM-2.5~\citep{das2023decoder}, Toto-1.0~\citep{cohen2025time}, Moirai-2.0~\citep{woo2024unified}, TabPFN-TS~\citep{hoo2025tables}, COSMIC~\citep{auer2025zero}, Sundial~\citep{liu2025sundial}, and Chronos-Bolt~\citep{ansari2024chronos}, the latest publicly released version of Chronos.
As additional baselines, we also included AutoARIMA, AutoETS, AutoTheta, and their ensemble~\citep{petropoulos2020simple}, representing well-established methods from the statistical forecasting literature~\citep{hyndman2018forecasting}.
We compare \ourmodel only with the aforementioned models and exclude task-specific deep learning models from our evaluation, as prior studies~\citep{aksu2024gift,ansari2024chronos} --- which include \gifteval and \chronosbenchii, two of the three benchmarks considered in our work --- have shown that pretrained models perform comparably to or better than task-specific models on average.

\begin{figure}[h]
    \centering
    \subfloat[]{\includegraphics[width=0.9\textwidth]{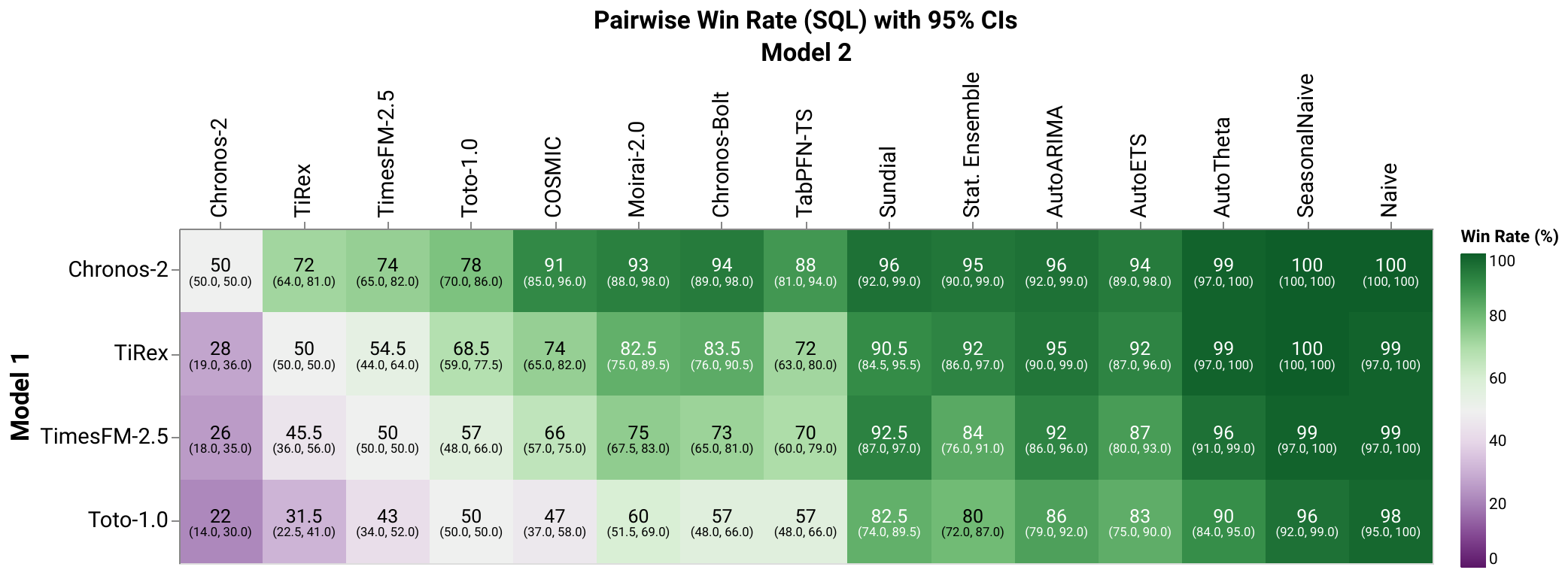}}
    \\
    \subfloat[]{\includegraphics[width=0.9\textwidth]{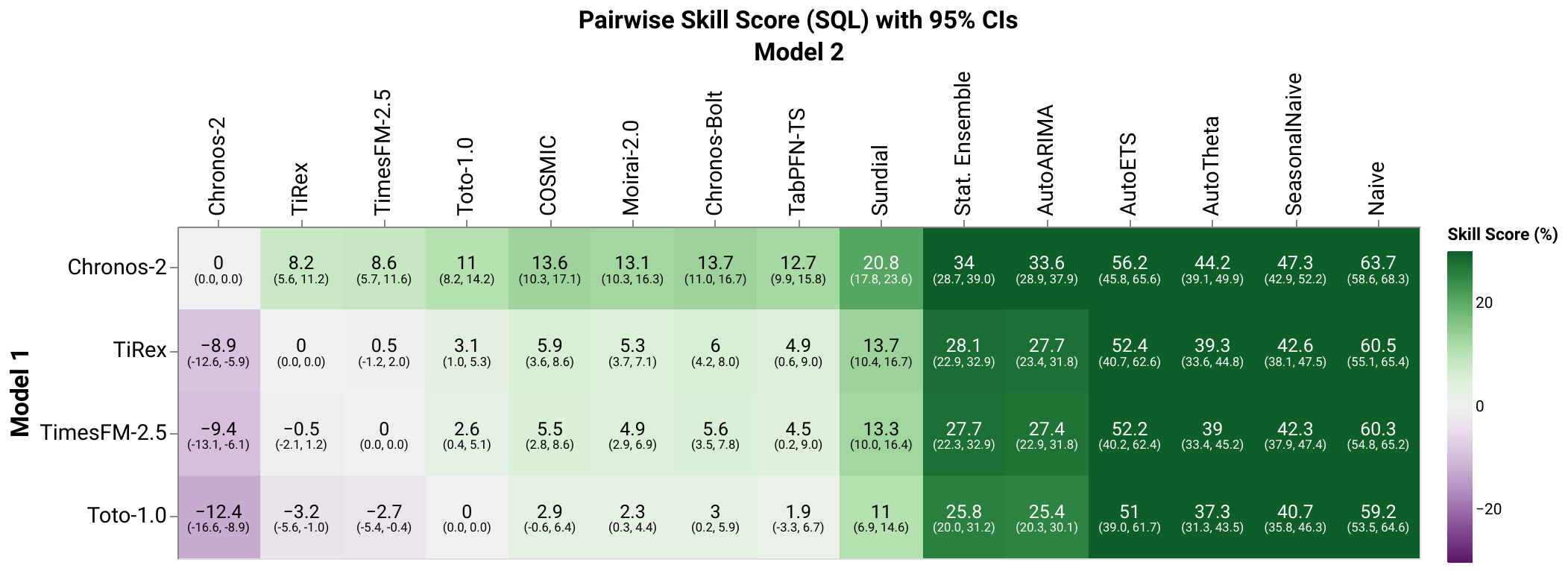}}
    \vspace{-1em}
    \caption{The pairwise win rates (a) and skill scores (b) of the top-4 pretrained models on \fevbench with 95\% confidence intervals (CIs) obtained through bootstrapping. \ourmodel outperforms the next best models (TiRex and TimesFM) by a statistically significant margin on both metrics. The complete plot and results for other forecasting metrics can be found in \Cref{fig:fev-pairwise-win-sql,fig:fev-pairwise-win-wql,fig:fev-pairwise-win-mase,fig:fev-pairwise-win-wape,fig:fev-pairwise-skill-sql,fig:fev-pairwise-skill-wql,fig:fev-pairwise-skill-mase,fig:fev-pairwise-skill-wape} (Appendix).} 
    \vspace{-1em}
    \label{fig:fev-pair-top-4-sql}
\end{figure}

Following \citet{shchur2025fev}, we report both average win rates ($W$) and skill scores ($S$) for all models. These metrics are mathematically equivalent to the average rank ($R$) and \emph{geometric mean relative error} (
$G$) metrics used in prior work~\citep{ansari2024chronos,aksu2024gift}. Specifically, $R = 1 + (1 - \frac{W}{100})(N - 1)$ and $G = 1 - \frac{S}{100}$, where $N$ is the number of evaluated models. However, win rates and skill scores provide more
interpretable summaries. The win rate measures the proportion of pairwise comparisons in which a model outperforms other models, while the skill score reflects the average percentage improvement over a baseline --- in our case,
the Seasonal Naive model. For a detailed discussion, we refer the reader to \citet{shchur2025fev}.

\begin{table}[h]
    \centering
    \subfloat[]{
    \resizebox{0.46\textwidth}{!}{
    \begin{tabular}{lrr}
\toprule
\textbf{Model} & \textbf{Avg. Win Rate (\%)} & \textbf{Skill Score (\%)} \\
\midrule
\rowcolor{AccentColorLight} Chronos-2 & \bfseries 81.9 & \bfseries 51.4 \\
TimesFM-2.5 & 77.5 & 51.0 \\
TiRex & 76.5 & 50.2 \\
Toto-1.0 & 67.4 & 48.6 \\
Moirai-2.0 & 64.4 & 48.4 \\
COSMIC & 56.4 & 44.5 \\
\rowcolor{AccentColorSuperLight} Chronos-Bolt & 53.8 & 42.6 \\
TabPFN-TS & 53.5 & 43.1 \\
Sundial & 49.1 & 44.1 \\
AutoARIMA & 21.8 & 8.8 \\
Seasonal Naive & 16.6 & 0.0 \\
AutoTheta & 16.0 & -24.4 \\
AutoETS & 15.2 & -648.9 \\
\bottomrule
\end{tabular}

    }
    }
    \quad
    \subfloat[]{
    \resizebox{0.46\textwidth}{!}{
    \begin{tabular}{lrr}
\toprule
\textbf{Model} & \textbf{Avg. Win Rate (\%)} & \textbf{Skill Score (\%)} \\
\midrule
\rowcolor{AccentColorLight} Chronos-2 & \bfseries 83.8 & \bfseries 30.2 \\
TimesFM-2.5 & 77.7 & 29.5 \\
TiRex & 71.9 & 27.6 \\
Moirai-2.0 & 64.3 & 27.2 \\
Toto-1.0 & 61.3 & 25.2 \\
\rowcolor{AccentColorSuperLight} Chronos-Bolt & 58.4 & 19.2 \\
Sundial & 53.4 & 25.0 \\
COSMIC & 51.9 & 20.8 \\
TabPFN-TS & 45.4 & 16.6 \\
AutoARIMA & 24.4 & -7.4 \\
AutoETS & 19.5 & -21.2 \\
Seasonal Naive & 19.4 & 0.0 \\
AutoTheta & 18.5 & -9.0 \\
\bottomrule
\end{tabular}

    }
    }
    
    \caption{\textbf{\textit{\gifteval results.}} The average win rate and skill score with respect to the (a) weighted quantile loss (WQL) and (b) mean absolute scaled error (MASE) metrics. Higher values are better for both. \ourmodel outperforms previous best models, TimesFM-2.5 and TiRex. Baseline results have been taken from the \gifteval leaderboard~\citep{aksu2024gift}.}
    \vspace{-1em}
    \label{tab:gift-results}
\end{table}

\paragraph{\fevbench.}
This benchmark consists of 100 forecasting tasks and offers the most comprehensive coverage of diverse real-world scenarios, including tasks with covariates. None of these datasets or tasks were seen by \ourmodel during training.
\Cref{tab:fev-results-sql} reports results on \fevbench with respect to the scaled quantile loss (SQL) metric which evaluates the probabilistic forecasting performance.
\ourmodel outperforms existing time series foundation models by a significant margin, both in win rate and skill score.
\fevbench also provides tooling to answer questions like: ``\emph{Does Model A outperform Model B in a statistically significant way?}''.
These pairwise comparisons with 95\% confidence intervals (CIs), shown in \Cref{fig:fev-pair-top-4-sql}, further confirm that \ourmodel surpasses the next best models (TiRex and TimesFM-2.5) by a statistically significant margin.
Specifically, the CIs of the pairwise win rates and skill scores of \ourmodel against any baseline do not include 50\% and 0\%, respectively.

\paragraph{\gifteval.}
The \gifteval benchmark comprises 97 tasks derived from 55 datasets, with a particular emphasis on high-frequency time series and long-horizon forecasting. The results in \Cref{tab:gift-results} show that \ourmodel surpasses the previously leading models (TiRex and TimesFM-2.5) in win rate and skill score under both the weighted quantile loss (WQL) and mean absolute scaled error (MASE) metrics. When constructing the pretraining corpus for \ourmodel, we carefully ensured that it did not overlap with the test portions of any \gifteval task at any sampling frequency. Nonetheless, the corpus does include partial overlap with the training portions of some \gifteval datasets. For strictly zero-shot results, we refer the reader to \Cref{sec:ablations}, where we evaluate a variant of \ourmodel trained exclusively on synthetic data.

\begin{table}[h]
    \centering
    \subfloat[]{
    \resizebox{0.46\textwidth}{!}{
    \begin{tabular}{lrr}
\toprule
\textbf{Model} & \textbf{Avg. Win Rate (\%)} & \textbf{Skill Score (\%)} \\
\midrule
\rowcolor{AccentColorLight} Chronos-2 & \bfseries 79.8 & \bfseries 46.6 \\
TiRex & 70.4 & 41.7 \\
TimesFM-2.5 & 70.0 & 42.4 \\
Toto-1.0 & 60.9 & 41.9 \\
Moirai-2.0 & 56.0 & 40.9 \\
\rowcolor{AccentColorSuperLight} Chronos-Bolt & 49.4 & 39.3 \\
TabPFN-TS & 46.3 & 32.6 \\
COSMIC & 42.8 & 36.7 \\
Sundial & 14.4 & 24.1 \\
Seasonal Naive & 10.1 & 0.0 \\
\bottomrule
\end{tabular}

    }
    }
    \quad
    \subfloat[]{
    \resizebox{0.46\textwidth}{!}{
    \begin{tabular}{lrr}
\toprule
\textbf{Model} & \textbf{Avg. Win Rate (\%)} & \textbf{Skill Score (\%)} \\
\midrule
\rowcolor{AccentColorLight} Chronos-2 & \bfseries 81.5 & \bfseries 26.5 \\
TimesFM-2.5 & 71.6 & 23.3 \\
TiRex & 67.1 & 22.2 \\
Toto-1.0 & 58.0 & 22.3 \\
Moirai-2.0 & 53.5 & 19.8 \\
\rowcolor{AccentColorSuperLight} Chronos-Bolt & 50.6 & 20.4 \\
COSMIC & 42.0 & 18.1 \\
TabPFN-TS & 40.1 & 10.5 \\
Sundial & 21.8 & 9.5 \\
Seasonal Naive & 13.8 & 0.0 \\
\bottomrule
\end{tabular}

    }
    }
    
    \caption{\textbf{\textit{\chronosbenchii results.}} The average win rate and skill score with respect to the (a) weighted quantile loss (WQL) and (b) mean absolute scaled error (MASE) metrics. Higher values are better for both. \ourmodel achieves the best results across all metrics.}
    \vspace{-1em}
    \label{tab:zs-results}
\end{table}

\paragraph{\chronosbenchii.}
Originally proposed in \citet{ansari2024chronos} to evaluate the first Chronos models, this benchmark comprises 27 tasks, the majority of which involve short histories (fewer than 300 time steps on average).
None of these datasets were included in the training corpus of \ourmodel.
On this benchmark, \ourmodel consistently outperforms existing models in terms of the win rate and skill score under both probabilistic (WQL) and point (MASE) forecasting metrics, as shown in \Cref{tab:zs-results}.

Taken together, these results show that \ourmodel not only outperforms all competing models across the three benchmarks but also substantially improves over Chronos-Bolt, its predecessor, highlighting the impact of the architectural and training improvements in \ourmodel.

\subsection{Improvements with In-context Learning}
\label{sec:icl-improvements}

The results in \Cref{sec:bench-results} correspond to \ourmodel with in-context learning (ICL) enabled, specifically in the \emph{full cross learning} mode described in \Cref{sec:inference}.
In this section, we disentangle the gains from ICL compared to univariate inference.
To this end, we split \texttt{fev-bench} into three subsets: the \emph{univariate subset} with 32 tasks involving a single target time series without covariates, the \emph{multivariate subset} with 26 tasks containing multiple targets but no covariates, and the \emph{covariates subset} with 42 tasks that include at least one past-only or known covariate.
We compare \ourmodel with ICL to its univariate inference mode on these three subsets, as well as on \gifteval and \chronosbenchii.
In the univariate mode, each time series in the batch is forecast independently, and covariates, if present, are ignored.

\begin{figure}[h]
    \centering
    \subfloat[]{\includegraphics[width=0.31\textwidth]{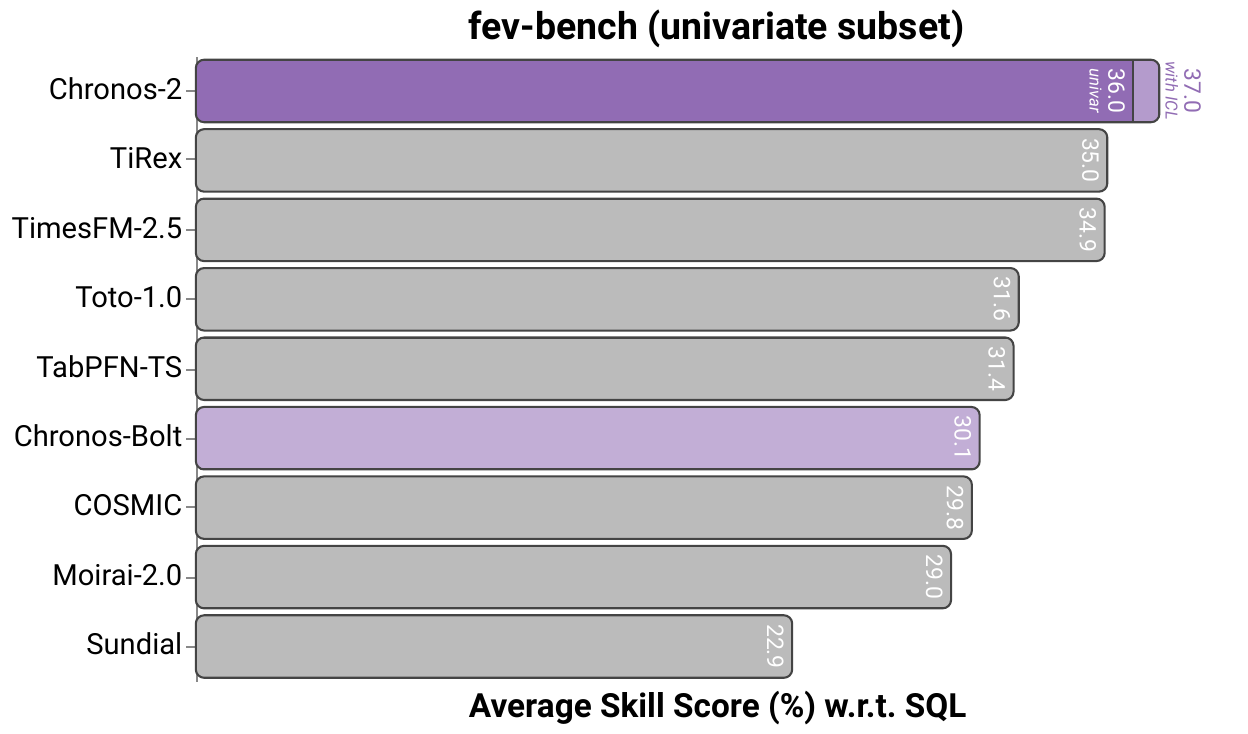}}
    \quad
    \subfloat[]{\includegraphics[width=0.31\textwidth]{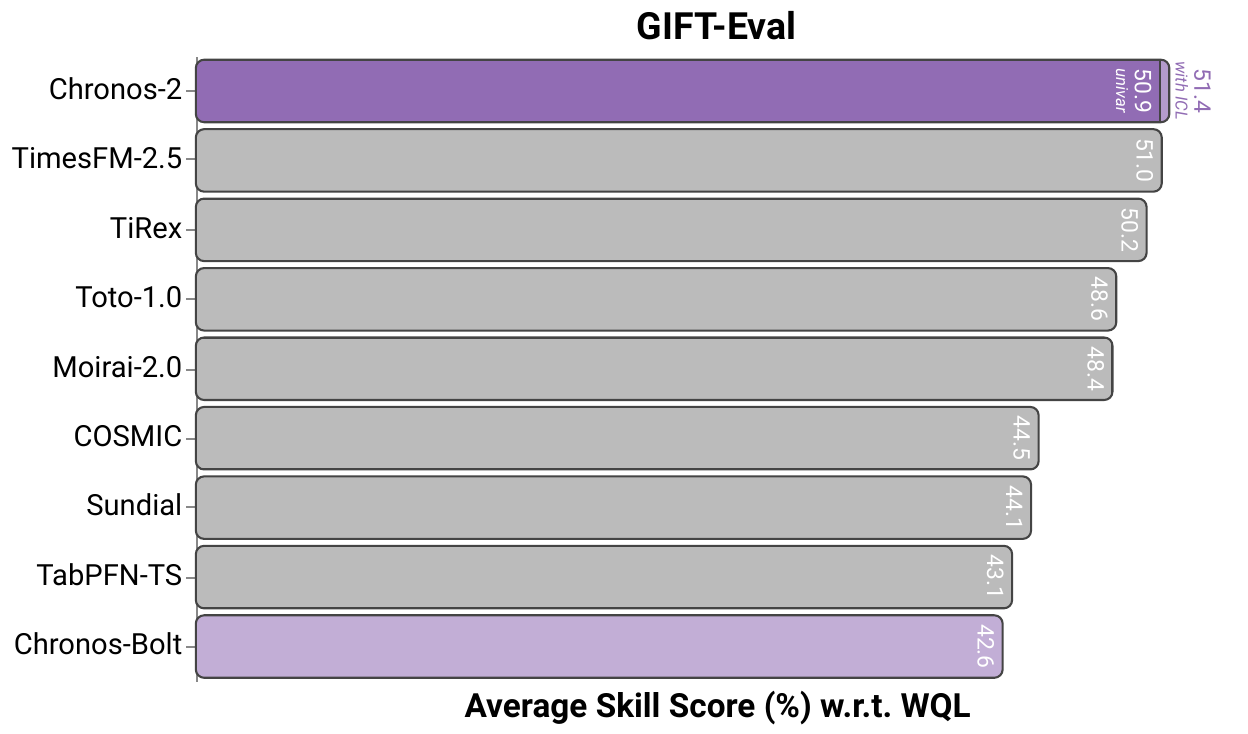}}
    \quad
    \subfloat[]{\includegraphics[width=0.31\textwidth]{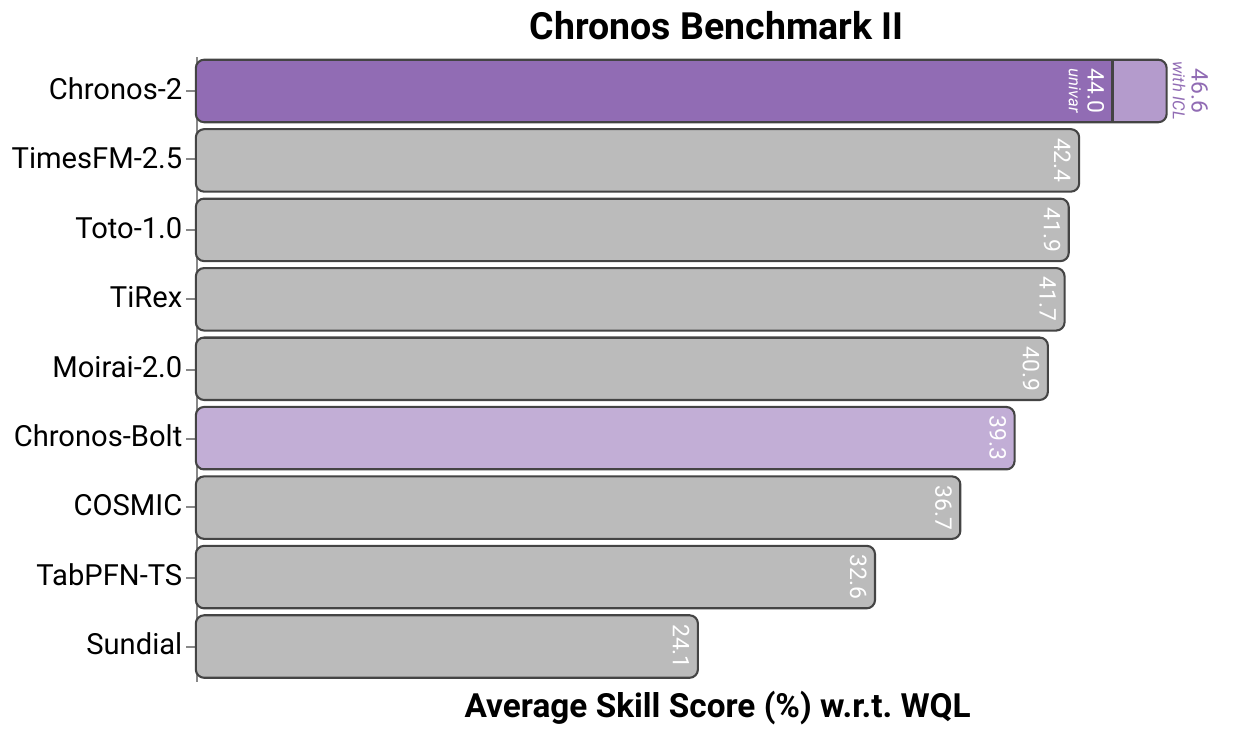}}
    \vspace{-1em}
    \caption{\ourmodel's probabilistic forecasting results in univariate mode and the corresponding improvements from in-context learning (ICL), shown as stacked bars on (a) the univariate subset of \fevbench, (b) \gifteval, and (c) \chronosbenchii. For these univariate benchmarks, ICL enables cross-learning, allowing the model to share information across items within a batch and thereby generate more accurate forecasts than univariate inference alone. Results for point forecasting metrics are available in \Cref{fig:univar-icl-improve-point} (Appendix).}
    \vspace{-1em}
    \label{fig:univar-icl-improve}
\end{figure}

\paragraph{Univariate Tasks.} ICL provides improvements in skill score on univariate tasks, as shown in \Cref{fig:univar-icl-improve}.
The effect is especially strong on \chronosbenchii (\Cref{fig:univar-icl-improve} (b)), which contains many tasks with short contexts.
This demonstrates that \ourmodel can leverage information from related time series to improve predictions when ICL is enabled, particularly when limited time series history is available.

\begin{figure}[h]
    \centering
    \subfloat[]{\includegraphics[width=0.48\textwidth]{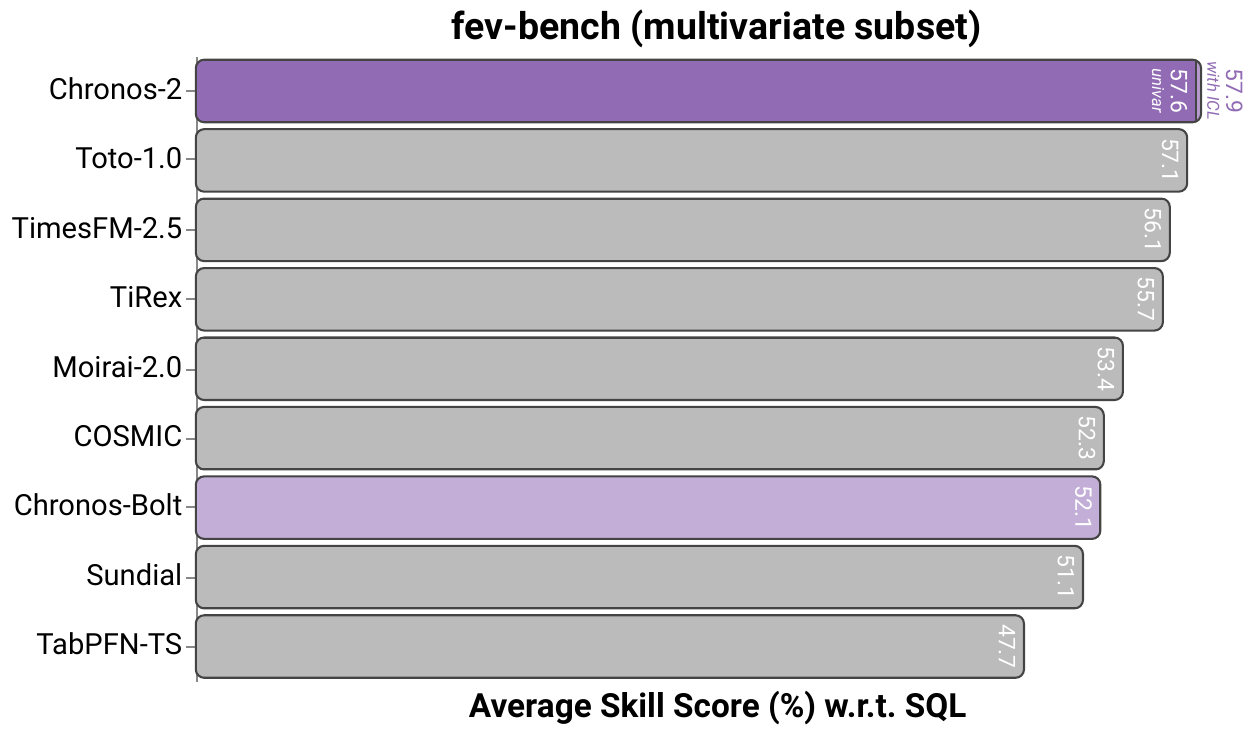}\label{fig:multi-icl-improve}}
    \quad
    \subfloat[]{\includegraphics[width=0.48\textwidth]{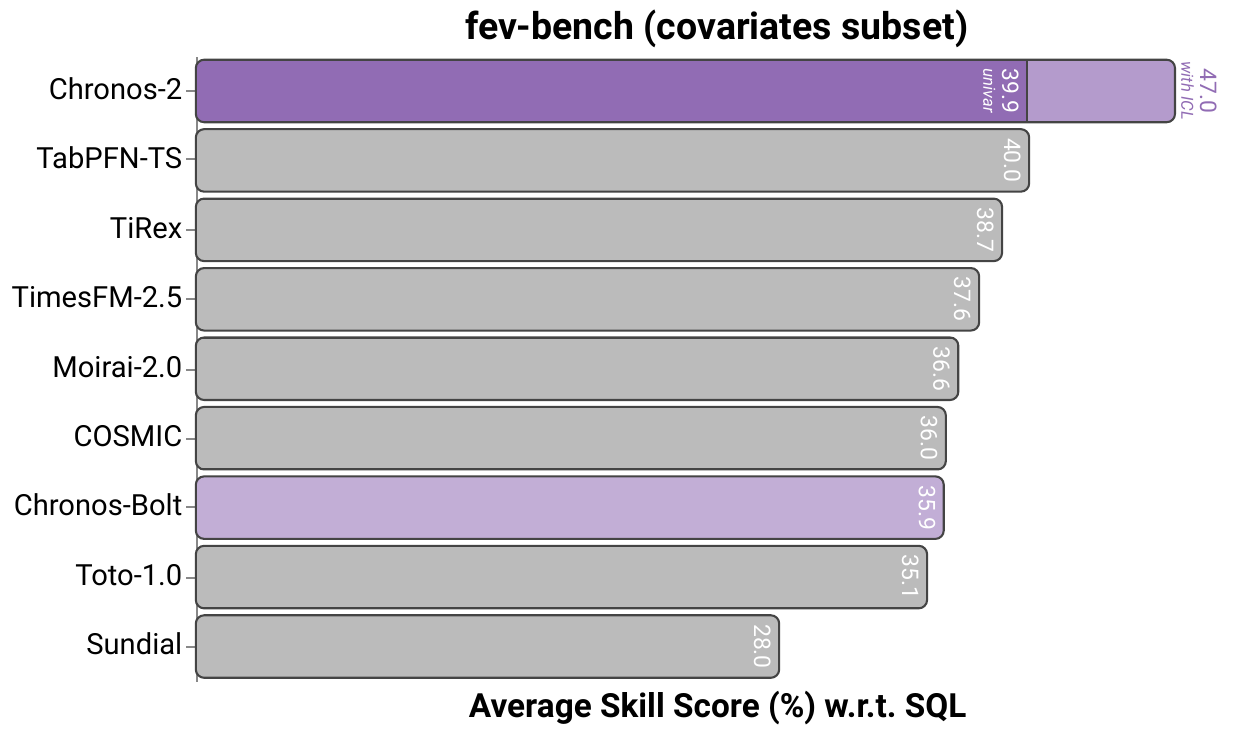}\label{fig:covariates-icl-improve}}
    \vspace{-1em}
    \caption{\ourmodel's probabilistic forecasting results in univariate mode and the corresponding gains from in-context learning (ICL), shown as stacked bars on the multivariate and covariates subsets of \fevbench. On multivariate tasks, ICL provides only modest improvements, though \ourmodel in univariate mode already surpasses the multivariate Toto-1.0 model. On the covariates subset, however, ICL delivers the largest gains, demonstrating \ourmodel's ability to effectively use covariates. Besides \ourmodel, only TabPFN-TS and COSMIC support covariates, and \ourmodel outperforms all baselines (including TabPFN-TS and COSMIC) by a wide margin. Results for point forecasting metrics are available in \Cref{fig:multi-icl-improve-point,fig:covariates-icl-improve-point} (Appendix).}
    \vspace{-1em}
\end{figure}

\paragraph{Multivariate Tasks.}
On the multivariate subset of \texttt{fev-bench}, ICL yields only modest gains over univariate inference (\Cref{fig:multi-icl-improve} (a)).
Interestingly, in univariate mode, \ourmodel even outperforms Toto-1.0, a model which natively supports multivariate forecasting.
This suggests that while these tasks involve multiple variates with potentially shared dynamics, the benefits of explicit multivariate modeling can be limited.
One possible intuition comes from Takens’s Embedding Theorem~\citep{takens2006detecting}, which implies that the dynamics of a system can often be reconstructed from delayed observations of a single variable.
In practice, this means that with sufficiently long histories, a strong univariate model may capture much of the same structure as a multivariate model.
Similar empirical findings have been reported elsewhere; for example, \citet{Nie2023PatchTST} observed that univariate (``channel-independent'') models often perform on par with multivariate (``channel-dependent'') models, albeit on a different benchmark.

\paragraph{Tasks with Covariates.}
The largest gains with ICL are observed on tasks with covariates (\Cref{fig:multi-icl-improve} (b)).
Here, the performance margin clearly demonstrates that \ourmodel with ICL can effectively exploit covariates to improve predictions compared to univariate inference, which ignores them.
Chronos-2 outperforms baselines by a large margin on this subset.
Unsurprisingly, the second spot is taken by TabPFN-TS, another model which supports (known) covariates.
These results underscore both the strength of \ourmodel and the limitations of existing pretrained models, most of which lack covariate support --- a capability of immense practical importance.

\begin{figure}[ht]
    \centering
    \subfloat[]{\includegraphics[width=0.48\textwidth]{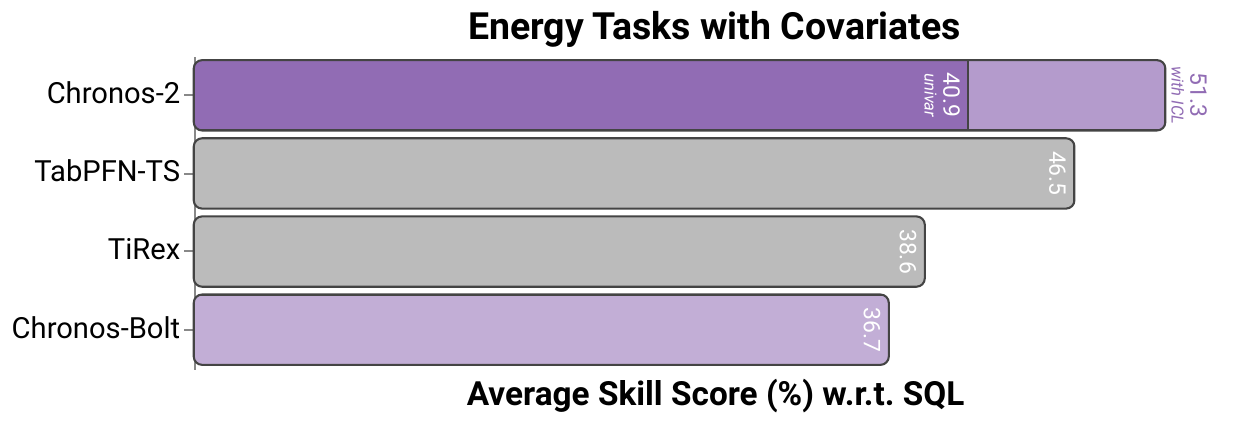}\label{fig:quant-energy}}
    \quad
    \subfloat[]{\includegraphics[width=0.48\textwidth]{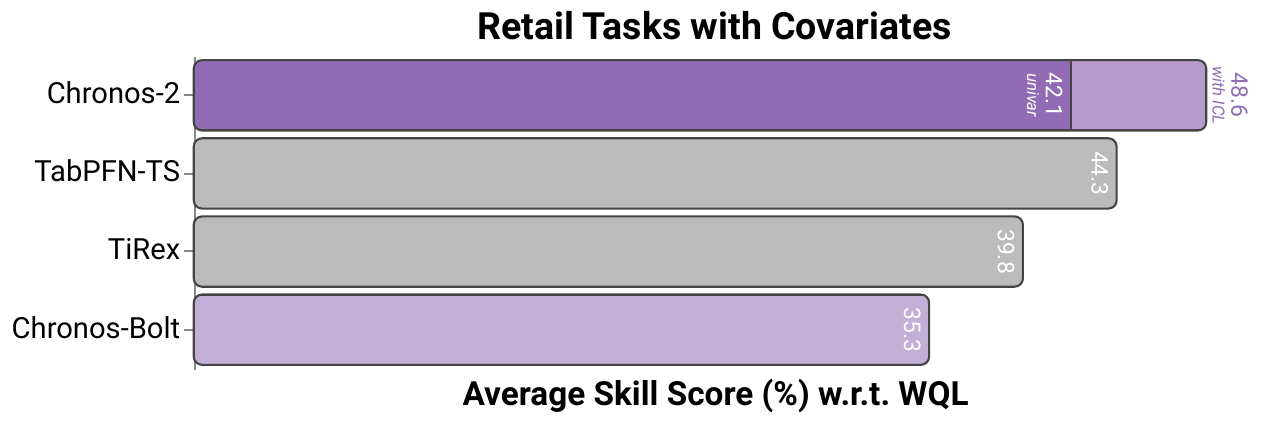}\label{fig:quant-retail}}
    \vspace{-0.5em}
    \caption{Comparison of \ourmodel against baselines on tasks which include dynamic covariates from the energy and retail domains. \ourmodel outperforms all baselines by a wide margin, including TabPFN-TS and TiRex, the strongest baselines on the covariates subset of \fevbench (\Cref{fig:covariates-icl-improve}). For retail, we consider the domain-appropriate WQL metric. Results for point forecasting metrics are available in \Cref{fig:quant-energy-point,fig:quant-retail-point} (Appendix).}
    \vspace{-1em}
\end{figure}

\begin{figure}[ht]
    \centering
    \includegraphics[width=0.8\linewidth]{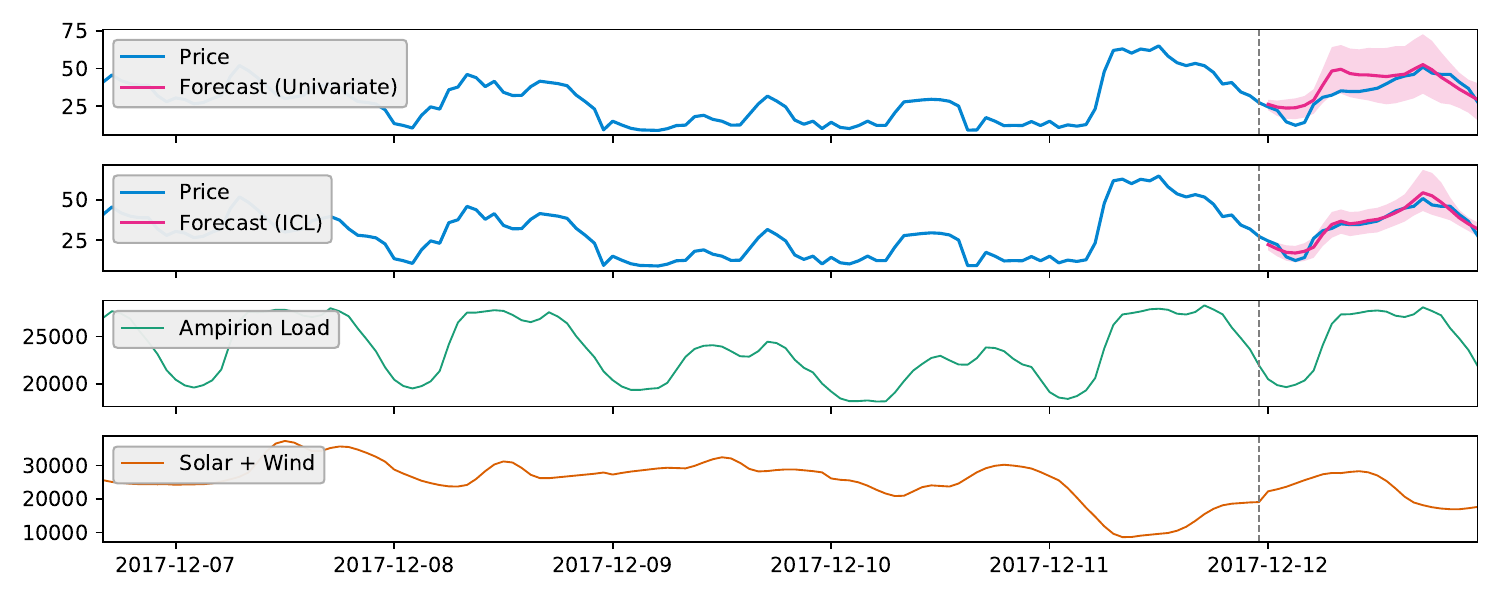}
    \vspace{-0.5em}
    \caption{Forecasts generated by \ourmodel in univariate mode (top), i.e., without covariates, and with in-context learning (second from top) on the energy price forecasting task. The dashed vertical gray line indicates the forecast start date and the shaded region represents 80\% prediction interval around the median forecast.  With ICL, \ourmodel leverages \texttt{Ampirion Load} and \texttt{Solar + Wind} covariates to produce a more accurate prediction.}
    \vspace{-1em}
    \label{fig:qual-energy}
\end{figure}
\begin{figure}[h]
    \centering
    \includegraphics[width=0.8\linewidth]{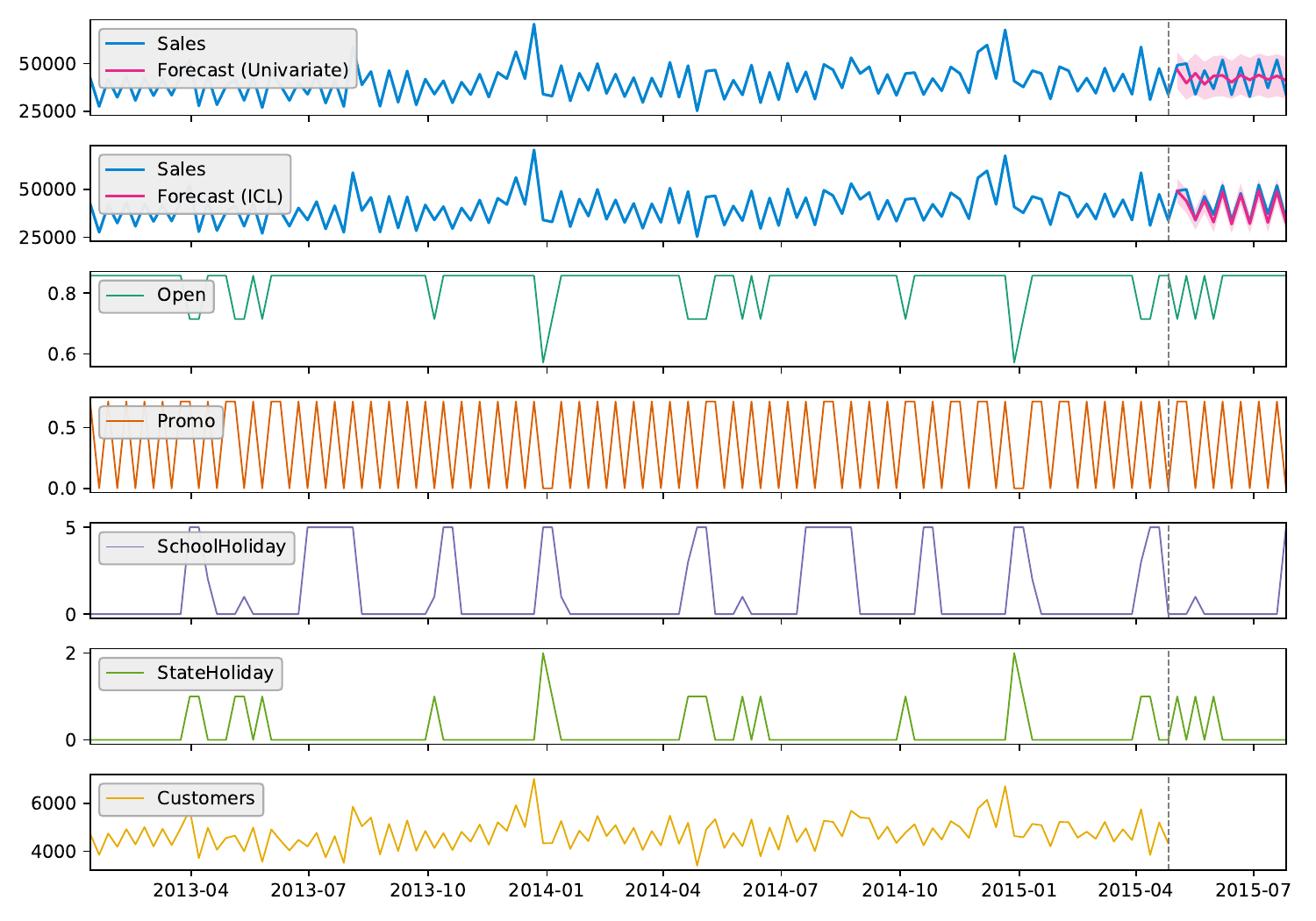}
    \vspace{-1em}
    \caption{Forecasts generated by \ourmodel in univariate mode (top), i.e., without covariates, and with in-context learning (second from top) on the Rossmann sales forecasting task. The dashed vertical gray line indicates the forecast start date and the shaded region represents 80\% prediction interval around the median forecast. With ICL, \ourmodel produces a substantially more accurate forecast by capturing the influence of promotion and holiday covariates on future sales.}
    \label{fig:qual-retail}
    \vspace{-1em}
\end{figure}

\subsection{Domain Case Studies}
\label{sec:domain}

We conducted further analysis on tasks from the \emph{energy} and \emph{retail} domains, where covariates often provide crucial information for accurate forecasting.
For both domains, we selected all tasks with dynamic covariates from \fevbench resulting in 16 and 17 tasks for energy and retail, respectively (see \Cref{tab:energy-tasks,tab:retail-tasks} in the Appendix for details).
As baselines, we used TabPFN-TS and TiRex, the two strongest models on the covariates subset of \fevbench, as shown in \Cref{fig:covariates-icl-improve}. The results in \Cref{fig:quant-energy,fig:quant-retail} demonstrate that \ourmodel consistently outperforms these baselines by a wide margin. Incorporating covariates provides a substantial boost in performance for \ourmodel, reinforcing their critical role in real-world forecasting tasks. Consistent with \Cref{fig:covariates-icl-improve}, the second-best results are achieved by TabPFN-TS, another model capable of leveraging covariates.

To illustrate how \ourmodel with ICL uses covariates, we compared forecasts produced in univariate mode versus with ICL.
We selected one task from each domain where ICL delivers the largest gains.

\Cref{fig:qual-energy} shows forecasts on the energy price forecasting task for Germany (EPF-DE), where the goal is to predict the hourly energy price for the next day using historical prices, day-ahead forecasts of the load and renewable (solar and wind) energy generation.
In the univariate mode, \ourmodel makes reasonable but imprecise predictions.
However, with ICL, \ourmodel effectively uses the covariates, producing significantly more accurate predictions.

The retail task in \Cref{fig:qual-retail} involves predicting next quarter's weekly store sales of Rossmann, a European drug store chain, using historical sales and covariates: historical customer footfall plus known covariates indicating store operation, promotion periods, and holidays.
\ourmodel's univariate forecast is nearly flat with high uncertainty.
In contrast, the ICL forecast leverages covariates --- particularly promotion and holiday information --- to capture the true sales dynamics over the forecast horizon.

\subsection{Ablation Studies}
\label{sec:ablations}

\begin{figure}[h]
    \centering
    \subfloat[]{\includegraphics[width=0.31\textwidth]{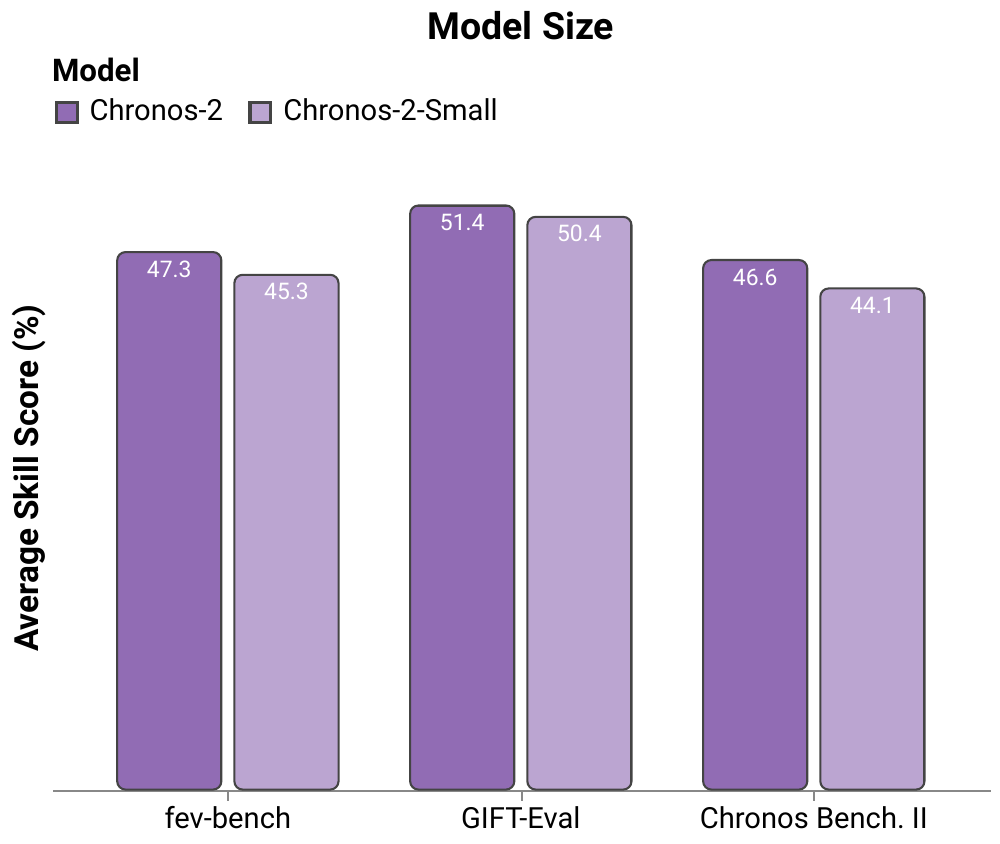}\label{fig:model-size}}
    \quad
    \subfloat[]{\includegraphics[width=0.31\textwidth]{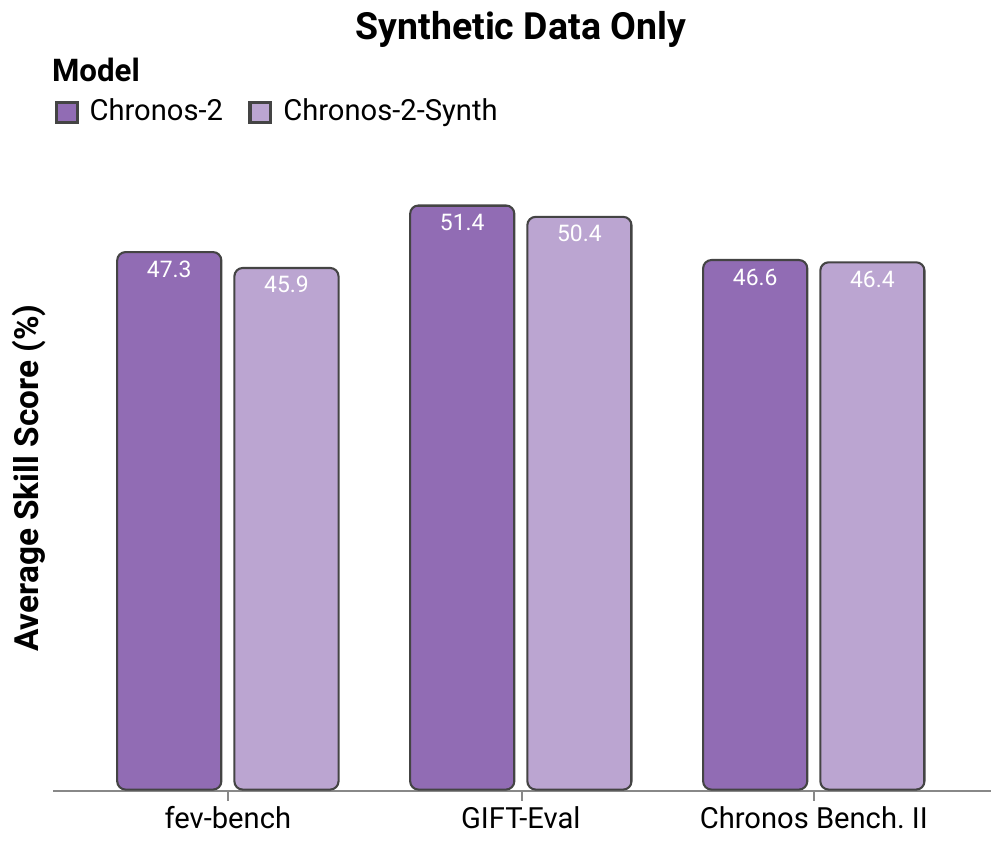}\label{fig:synth-only}}
    \quad
    \subfloat[]{\includegraphics[width=0.31\textwidth]{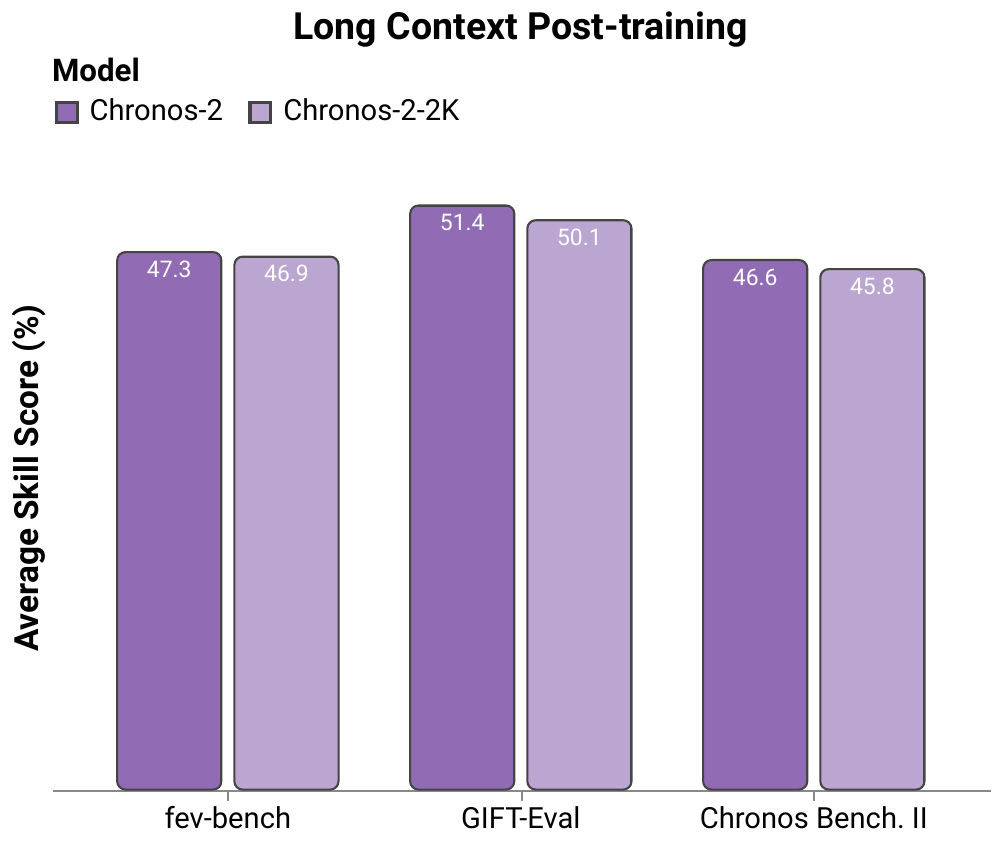}\label{fig:long-context}}
    \vspace{-1em}
    \caption{Comparison of the main \ourmodel model (120M parameters) with (a) a smaller 28M-parameter model, (b) a model trained exclusively on synthetic data, and (c) the main model prior to long-context post-training.}
    \vspace{-1em}
    \label{fig:analysis}
\end{figure}

In this section, we present additional experiments and ablations that disentangle the impact of different design choices. We investigate the performance of \ourmodel across different parameter counts, evaluate models trained exclusively on synthetic data, and demonstrate the importance of post-training on long-context scenarios.

\paragraph{Model Size.} 
We trained a \emph{small} model with 28M parameters to understand the impact of model size on forecasting performance.
As shown in \Cref{fig:model-size}, the small model delivers strong performance despite its reduced size.
On \gifteval, for instance, its skill score lags the base model by as little as 1\% points, while offering nearly 2$\times$ faster inference.
This makes it particularly suitable for low-resource environments, such as CPU-only settings, or applications where inference speed is prioritized over maximum forecast accuracy.

\paragraph{Synthetic Data Only.}
Synthetic time series data has played a pivotal role in advancing pretrained forecasting models~\citep{ansari2024chronos,das2023decoder}.
TabPFN-TS~\citep{hoo2025tables} demonstrated that strong performance is achievable even when training relies exclusively on synthetic data.
To examine the limits of this approach, we trained a version of \ourmodel using only synthetic data.
On \chronosbenchii and \gifteval, this model (Chronos-2-Synth) performs only slightly below the version with real data in its pretraining corpus (\Cref{fig:synth-only}).
It also delivers strong results on \fevbench, though with a larger performance gap.
These results underscore the importance of synthetic data, suggesting that with further research, real data may not even be required for effective pretraining.

\paragraph{Long context Post-training.}
As described in \Cref{sec:training}, \ourmodel is initially trained with a context length of 2,048 time steps and then post-trained with an extended context of 8,192 steps. \Cref{fig:long-context} compares the base model (denoted \ourmodel-2K) with the post-trained variant. Extending the context length yields gains, particularly on the \gifteval benchmark, which contains many high-frequency datasets with long seasonal periods.

\section{Discussion}
\label{sec:discussion}

We introduced \ourmodel, a pretrained time series model designed to handle a wide range of forecasting scenarios --- including univariate, multivariate, and covariate-informed tasks --- in a zero-shot manner. Across three comprehensive benchmarks, \ourmodel consistently outperforms existing foundation models, demonstrating that in-context learning enhances forecasting performance across diverse task types.

A particularly large performance gap appears on covariate-informed tasks, where \ourmodel substantially surpasses prior foundation models. This highlights both the limitations of existing models and the critical role contextual information (e.g., covariates) plays in accurate forecasting. While \ourmodel supports only numeric and categorical covariates, extending pretrained models to incorporate multimodal inputs, such as text, represents a promising direction for future research~\citep{zhang2025does}. 

Our results further emphasize the importance of synthetic data in enabling generalist forecasting. The abilities of \ourmodel beyond univariate forecasting rely entirely on synthetic data, and ablation studies show that models trained solely on synthetic data perform only slightly worse than those trained on a mixture of real and synthetic datasets. We expect synthetic data to play an increasingly central role in advancing pretrained time series models. 

Finally, the flexible group attention mechanism in \ourmodel opens opportunities for further applications. For instance, time series could be grouped using sparse metadata or dense embeddings to enable \emph{retrieval-augmented forecasting}, potentially improving performance in small-data or cold-start scenarios.

\section*{Acknowledgements}
We thank the developers of open-source libraries used in the development of \ourmodel, including but not limited to \texttt{torch}~\citep{paszke2019pytorch}, \texttt{numpy}~\citep{harris2020array}, \texttt{pandas}~\citep{reback2020pandas,mckinney-proc-scipy-2010}, \texttt{statsmodels}~\citep{seabold2010statsmodels}, \texttt{transformers}~\citep{wolf-etal-2020-transformers}, \texttt{gluonts}~\citep{alexandrov2020gluonts}, \texttt{autogluon}~\citep{shchur2023autogluon}, \texttt{statsforecast}~\citep{garza2022statsforecast}, \texttt{einops}~\citep{rogozhnikov2022einops} and \texttt{scikit-learn}~\citep{pedregosa2011scikit}.
We also thank our colleagues at Amazon for their invaluable support in releasing \ourmodel: Kevin Ormiston, Jenna Larson, Larry Hardesty, Divya Sukumar, Lahari Chowtoori and Henri Yandell.
Finally, we are grateful to our fellow researchers for insightful discussions and their contributions to the field: Andrew Gordon Wilson, Michael Mahoney, Dmitry Efimov, Christoph Bergmeir, Valentin Flunkert, David Salinas, Imry Kissos, Devamanyu Hazarika, Tim Januschowski, Jan Gasthaus, William Gilpin, Annan Yu, Zelin He, Kashif Rasul, Rajat Sen, Yichen Zhou, Chenghao Liu, Taha Aksu, Gerald Woo, Emaad Khwaja and Ben Cohen.

\bibliography{main}
\bibliographystyle{tmlr}

\clearpage
\appendix
\section{Training Data}

\begin{table}[ht]
    \centering
    \resizebox{\textwidth}{!}{%
    \begin{tabular}{llrll}
    \toprule
    \textbf{Dataset Name} & \textbf{Frequencies} & \textbf{\# Time Series} & \textbf{Domain} & \textbf{Source} \\
    \midrule
    Electricity & 15min, 1H, 1W, 1D & 370 & Energy & \citet{godahewa2021monash} \\
    KDD Cup (2018) & 1H, 1D & 270 & Nature & \citet{godahewa2021monash} \\
    M4 (Daily) & 1D & 4227 & Various & \citet{makridakis2020m4} \\
    M4 (Hourly) & 1H & 414 & Various & \citet{makridakis2020m4} \\
    M4 (Monthly) & 1M & 48000 & Various & \citet{makridakis2020m4} \\
    M4 (Weekly) & 1W & 359 & Various & \citet{makridakis2020m4} \\
    Mexico City Bikes & 1H, 1D, 1W & 494 & Transport & \citet{ansari2024chronos} \\
    Pedestrian Counts & 1H, 1D, 1W & 66 & Transport & \citet{godahewa2021monash} \\
    Solar & 5min, 10min, 1H & 5166 & Energy & \citet{ansari2024chronos} \\
    Taxi & 30min, 1H & 2428 & Transport & \citet{salinas2019high} \\
    Uber TLC & 1H, 1D & 262 & Transport & \citet{fivethirtyeight_uber_tlc_foil_response} \\
    USHCN & 1D, 1W & 225280 & Nature & \citet{ansari2024chronos} \\
    Weatherbench & 1H, 1D, 1W & 225280 & Nature & \citet{rasp2020weatherbench} \\
    Wiki & 1H, 1D, 1W & 100000 & Web & \citet{ansari2024chronos} \\
    Wind Farms & 1H, 1D & 337 & Energy & \citet{godahewa2021monash} \\
    Temperature-Rain & 1D & 32072 & Nature & \citet{godahewa2021monash} \\
    London Smart Meters & 30min, 1D & 5560 & Energy & \citet{godahewa2021monash} \\
    Alibaba Cluster Trace (2018) & 5min, 1H & 100000 & Cloud Ops & \citet{woo2023pushing} \\
    Azure VM Traces (2017) & 5min, 1H & 100000 & Cloud Ops & \citet{woo2023pushing} \\
    Borg Cluster Data (2011) & 5min, 1H & 100000 & Cloud Ops & \citet{woo2023pushing} \\
    LargeST (2017) & 1H, 1D & 8196 & Transport & \citet{liu2023largest} \\
    Q-Traffic & 15min, 1H & 45148 & Transport & \citet{jiang2023libcity} \\
    Buildings 900K & 1H, 1D & 100000 & Energy & \citet{emami2023buildingsbench} \\
    \bottomrule
    \end{tabular}%
    }
    \caption{Real univariate datasets used for pretraining \ourmodel.}
    \label{tab:real-uni-datasets}
\end{table}

\section{Additional Results}
\label{app:additional-results}

\begin{table}[h]
    \centering
    \resizebox{\textwidth}{!}{
    \begin{tabular}{lrrrrr}
\toprule
\textbf{Model} & \textbf{Avg. Win Rate (\%)} & \textbf{Skill Score (\%)} & \textbf{Median runtime (s)} & \textbf{Leakage (\%)} & \textbf{\#Failures} \\
\midrule
\rowcolor{AccentColorLight} Chronos-2 & \bfseries 87.9 & \bfseries 35.5 & 3.6 & 0 & 0 \\
TiRex & 75.1 & 30.0 & 1.4 & 1 & 0 \\
TimesFM-2.5 & 74.4 & 30.3 & 16.9 & 8 & 0 \\
Toto-1.0 & 64.3 & 28.2 & 90.7 & 8 & 0 \\
Moirai-2.0 & 58.7 & 27.3 & 2.5 & 28 & 0 \\
COSMIC & 58.6 & 25.7 & 34.4 & 0 & 0 \\
\rowcolor{AccentColorSuperLight} Chronos-Bolt & 57.9 & 26.5 & 1.0 & 0 & 0 \\
TabPFN-TS & 55.7 & 27.6 & 305.5 & 0 & 2 \\
Sundial & 49.8 & 24.7 & 35.6 & 1 & 0 \\
Stat. Ensemble & 44.2 & 15.7 & 690.6 & 0 & 11 \\
AutoARIMA & 32.1 & 11.2 & 186.8 & 0 & 10 \\
AutoTheta & 30.3 & 11.0 & 9.3 & 0 & 0 \\
AutoETS & 30.2 & 2.3 & 17.0 & 0 & 3 \\
SeasonalNaive & 16.7 & 0.0 & 2.3 & 0 & 0 \\
Naive & 14.0 & -16.7 & 2.2 & 0 & 0 \\
\bottomrule
\end{tabular}

    }
    \caption{\textbf{\textit{\fevbench results.}} The average win rate and skill score are computed with respect to the mean absolute scaled error (MASE) metric on \fevbench. Higher values are better for both.}
    \label{tab:fev-results-mase}
\end{table}

\begin{table}[h]
    \centering
    \resizebox{\textwidth}{!}{
    \begin{tabular}{lrrrrr}
\toprule
\textbf{Model} & \textbf{Avg. Win Rate (\%)} & \textbf{Skill Score (\%)} & \textbf{Median runtime (s)} & \textbf{Leakage (\%)} & \textbf{\#Failures} \\
\midrule
\rowcolor{AccentColorLight} Chronos-2 & \bfseries 88.5 & \bfseries 51.5 & 3.6 & 0 & 0 \\
TiRex & 79.0 & 46.7 & 1.4 & 1 & 0 \\
TimesFM-2.5 & 76.8 & 46.8 & 16.9 & 8 & 0 \\
Toto-1.0 & 67.6 & 45.0 & 90.7 & 8 & 0 \\
COSMIC & 65.2 & 43.7 & 34.4 & 0 & 0 \\
TabPFN-TS & 64.8 & 45.8 & 305.5 & 0 & 2 \\
Moirai-2.0 & 62.8 & 43.9 & 2.5 & 28 & 0 \\
\rowcolor{AccentColorSuperLight} Chronos-Bolt & 60.5 & 43.2 & 1.0 & 0 & 0 \\
Sundial & 41.9 & 37.4 & 35.6 & 1 & 0 \\
Stat. Ensemble & 38.3 & 21.8 & 690.6 & 0 & 11 \\
AutoARIMA & 34.6 & 23.4 & 186.8 & 0 & 10 \\
AutoETS & 26.8 & -27.0 & 17.0 & 0 & 3 \\
AutoTheta & 21.3 & 7.8 & 9.3 & 0 & 0 \\
SeasonalNaive & 14.1 & 0.0 & 2.3 & 0 & 0 \\
Naive & 7.8 & -39.1 & 2.2 & 0 & 0 \\
\bottomrule
\end{tabular}

    }
    \caption{\textbf{\textit{\fevbench results.}} The average win rate and skill score are computed with respect to the weighted quantile loss (WQL) metric on \fevbench. Higher values are better for both.}
    \label{tab:fev-results-wql}
\end{table}

\begin{table}[h]
    \centering
    \resizebox{\textwidth}{!}{
    \begin{tabular}{lrrrrr}
\toprule
\textbf{Model} & \textbf{Avg. Win Rate (\%)} & \textbf{Skill Score (\%)} & \textbf{Median runtime (s)} & \textbf{Leakage (\%)} & \textbf{\#Failures} \\
\midrule
\rowcolor{AccentColorLight} Chronos-2 & \bfseries 85.4 & \bfseries 39.4 & 3.6 & 0 & 0 \\
TimesFM-2.5 & 74.1 & 33.8 & 16.9 & 8 & 0 \\
TiRex & 73.7 & 33.6 & 1.4 & 1 & 0 \\
Toto-1.0 & 65.1 & 31.5 & 90.7 & 8 & 0 \\
TabPFN-TS & 61.5 & 33.4 & 305.5 & 0 & 2 \\
COSMIC & 60.5 & 30.1 & 34.4 & 0 & 0 \\
Moirai-2.0 & 59.6 & 30.7 & 2.5 & 28 & 0 \\
\rowcolor{AccentColorSuperLight} Chronos-Bolt & 58.0 & 29.8 & 1.0 & 0 & 0 \\
Sundial & 47.7 & 27.3 & 35.6 & 1 & 0 \\
Stat. Ensemble & 43.0 & 17.7 & 690.6 & 0 & 11 \\
AutoETS & 30.8 & 4.3 & 17.0 & 0 & 3 \\
AutoARIMA & 30.8 & 13.3 & 186.8 & 0 & 10 \\
AutoTheta & 27.2 & 13.8 & 9.3 & 0 & 0 \\
Naive & 17.5 & -6.1 & 2.2 & 0 & 0 \\
SeasonalNaive & 15.2 & 0.0 & 2.3 & 0 & 0 \\
\bottomrule
\end{tabular}

    }
    \caption{\textbf{\textit{\fevbench results.}} The average win rate and skill score are computed with respect to the weighted absolute percentage error (WAPE) metric on \fevbench. Higher values are better for both.}
    \label{tab:fev-results-wape}
\end{table}

\begin{figure}[h]
    \centering
    \subfloat[]{\includegraphics[width=0.31\textwidth]{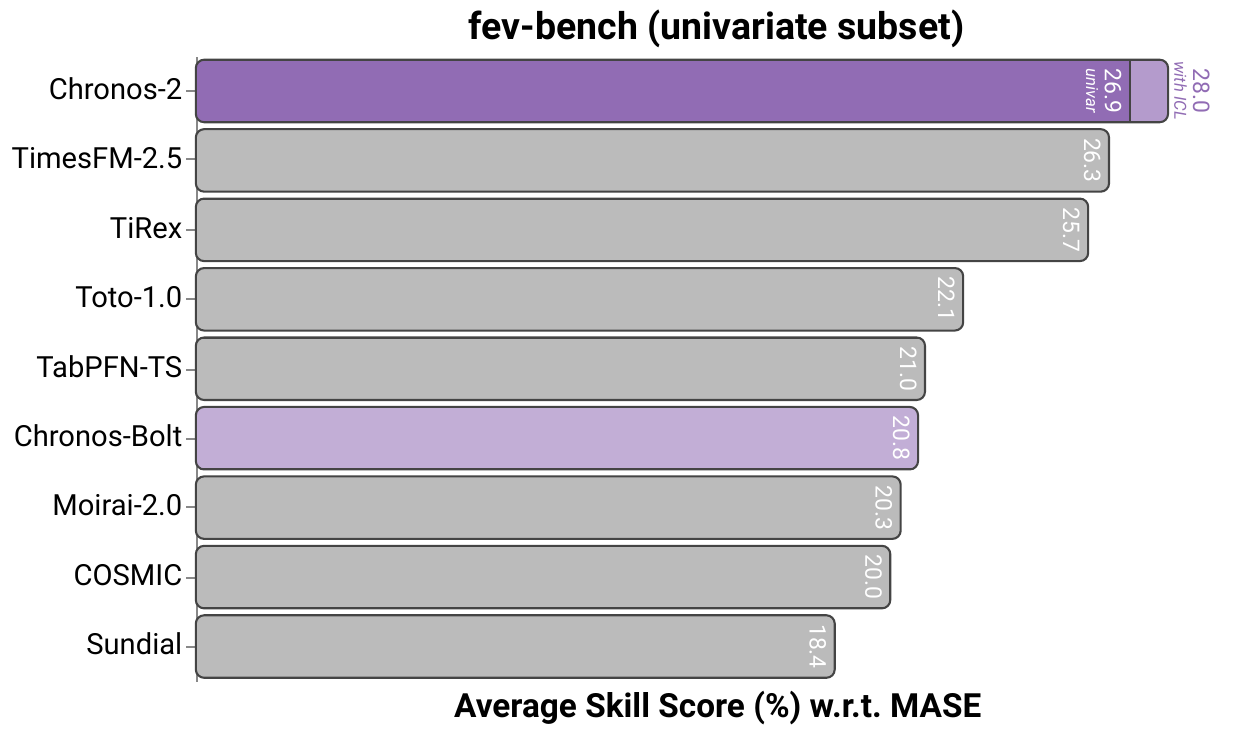}}
    \quad
    {\includegraphics[width=0.31\textwidth]{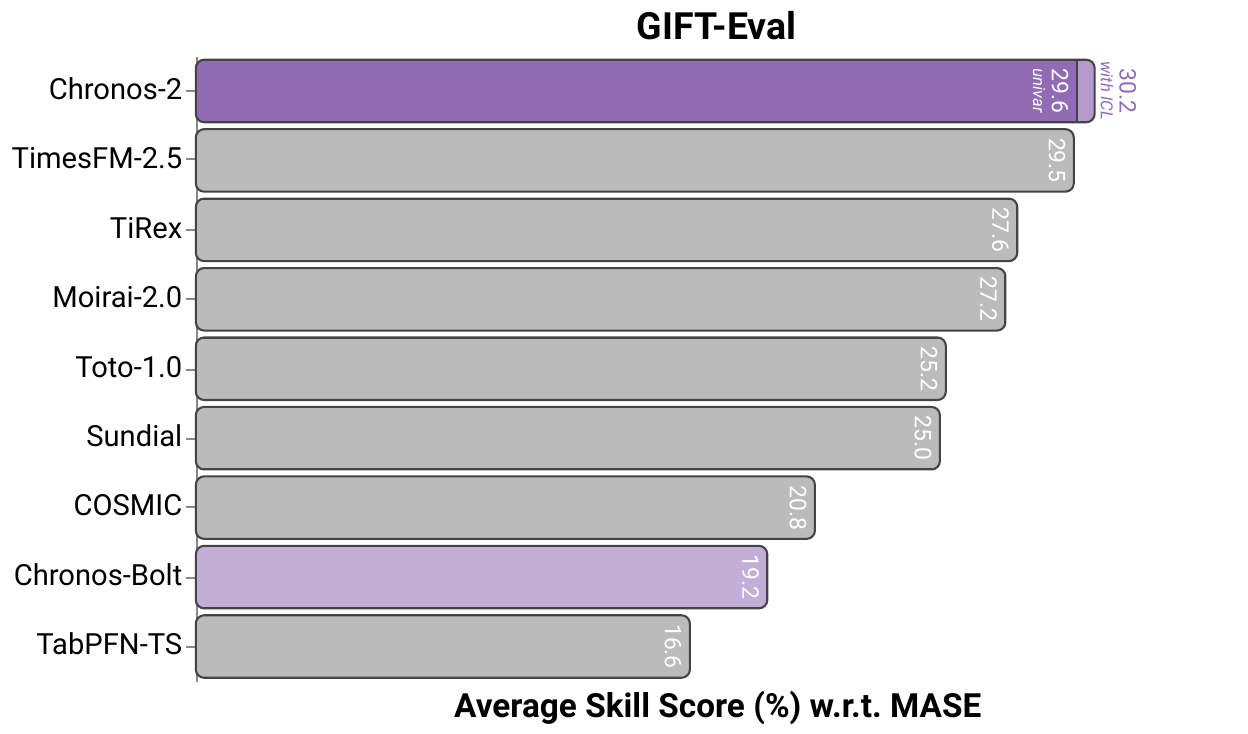}}
    \quad
    \subfloat[]{\includegraphics[width=0.31\textwidth]{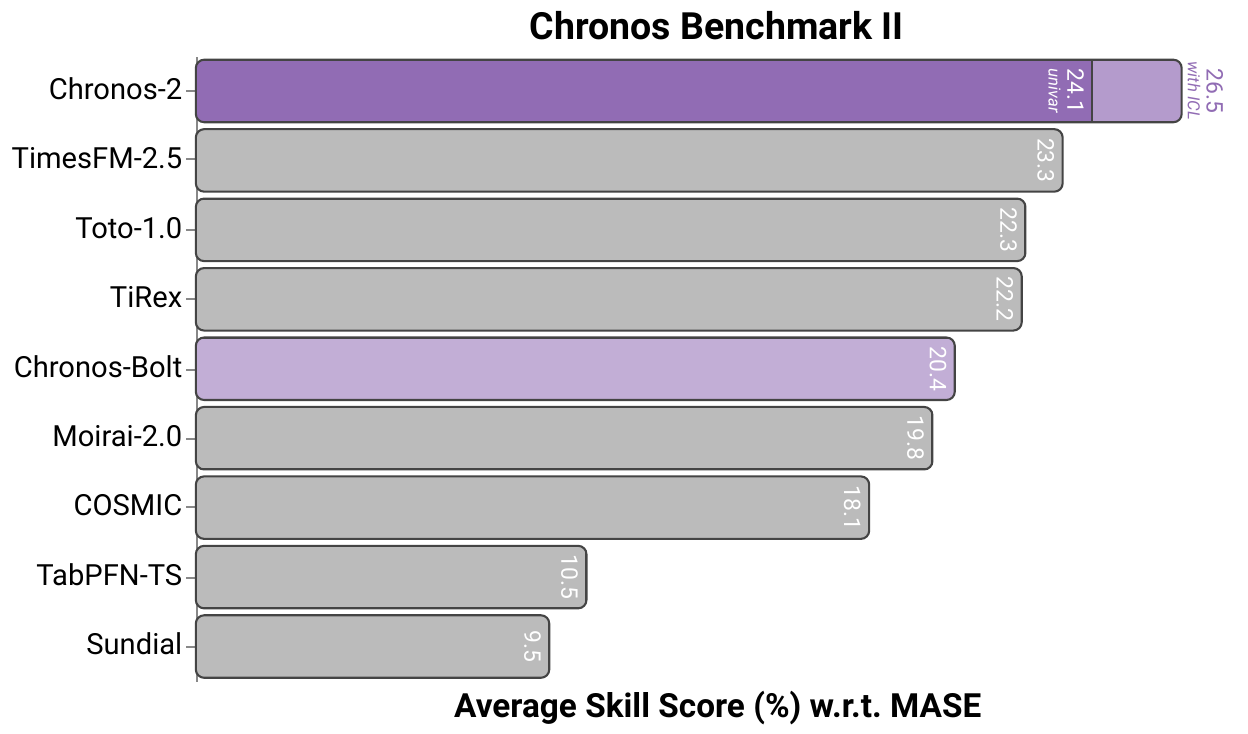}}
    \vspace{-1em}
    \caption{\ourmodel's point forecasting results in univariate mode and the corresponding improvements from in-context learning (ICL), shown as stacked bars on (a) the univariate subset of \fevbench, (b) \gifteval, and (c) \chronosbenchii.}
    \vspace{-1em}
    \label{fig:univar-icl-improve-point}
\end{figure}

\begin{figure}[h]
    \centering
    \subfloat[]{\includegraphics[width=0.48\textwidth]{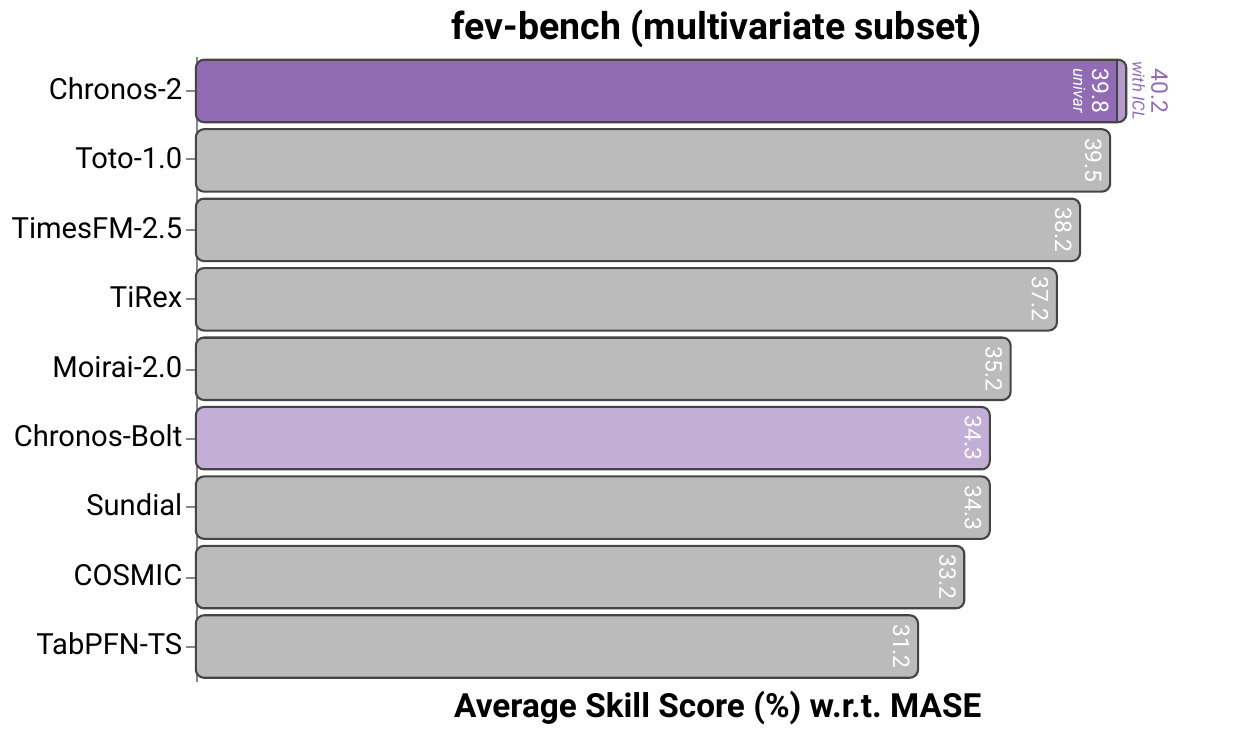}\label{fig:multi-icl-improve-point}}
    \quad
    \subfloat[]{\includegraphics[width=0.48\textwidth]{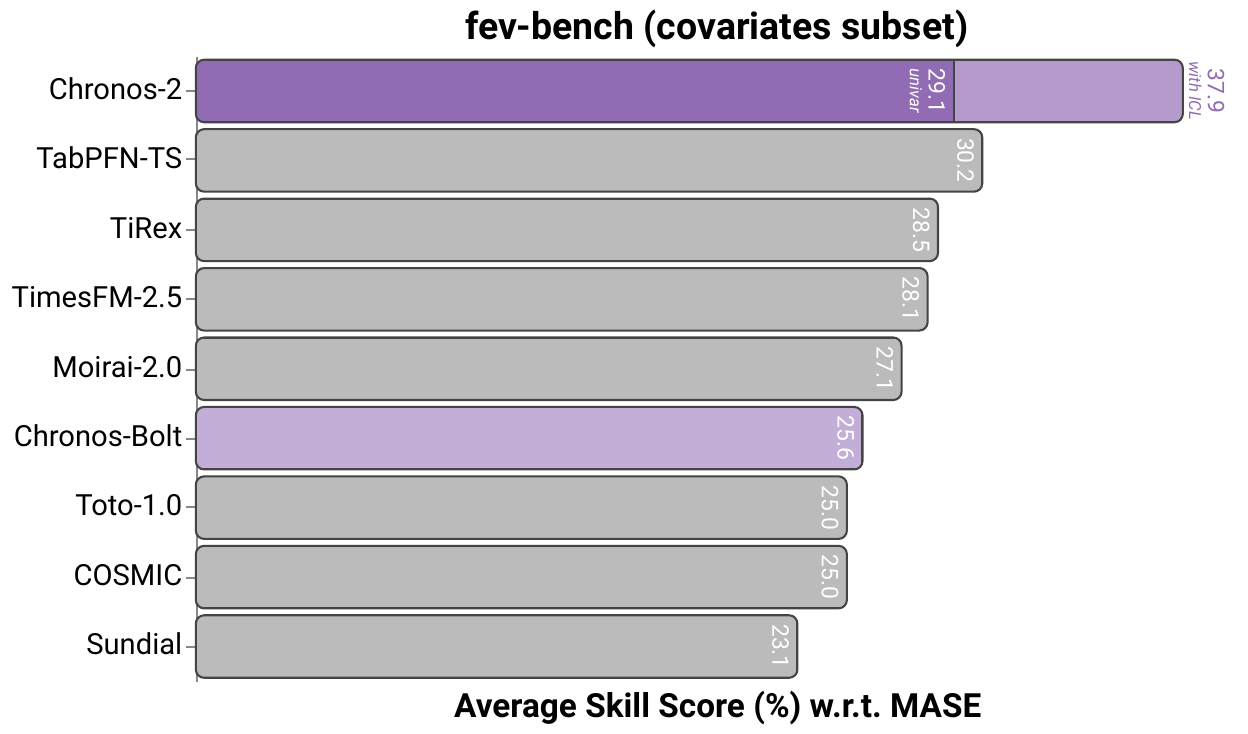}\label{fig:covariates-icl-improve-point}}
    \vspace{-1em}
    \caption{\ourmodel's point forecasting results in univariate mode and the corresponding gains from in-context learning (ICL), shown as stacked bars on the multivariate and covariates subsets of \fevbench.}
    \vspace{-1em}
\end{figure}

\begin{figure}[ht]
    \centering
    \subfloat[]{\includegraphics[width=0.48\textwidth]{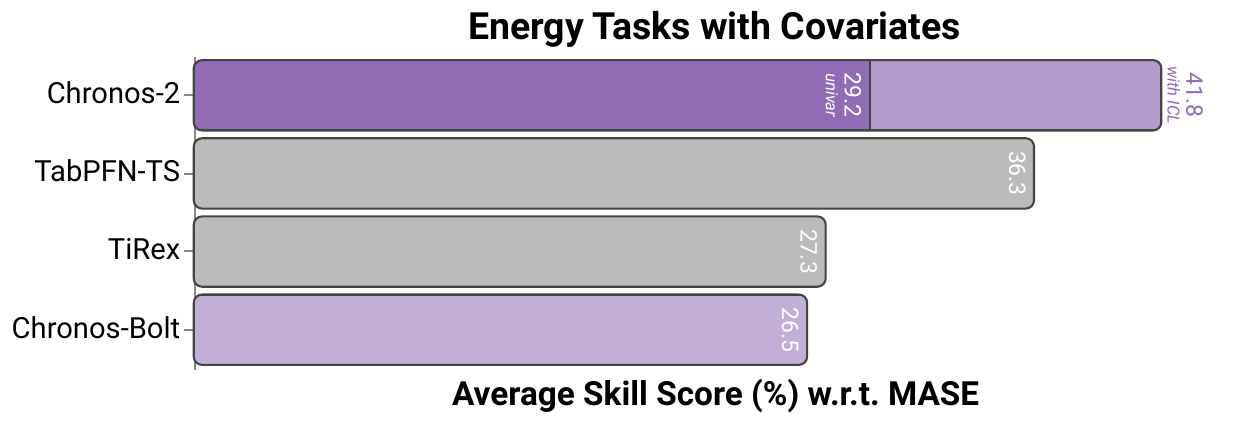}\label{fig:quant-energy-point}}
    \quad
    \subfloat[]{\includegraphics[width=0.48\textwidth]{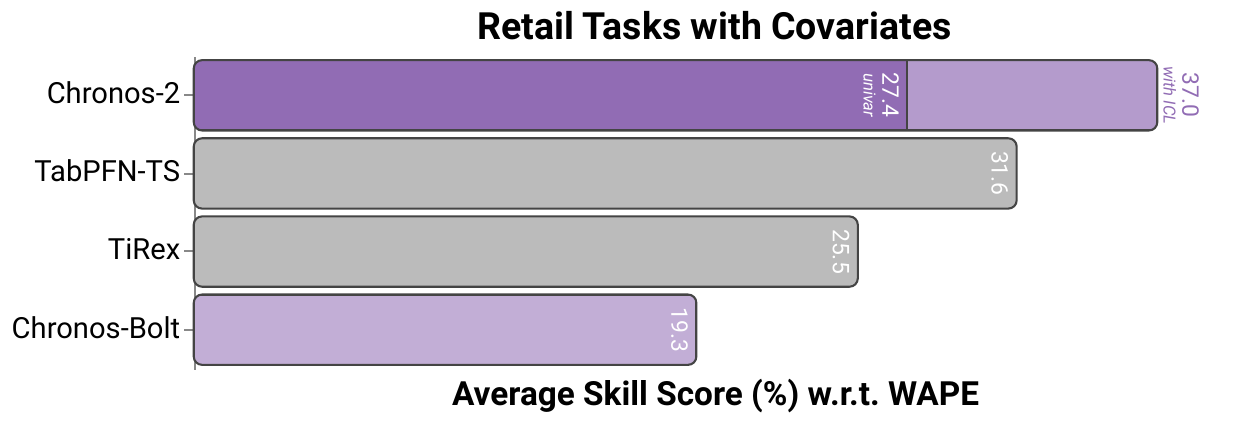}\label{fig:quant-retail-point}}
    \vspace{-0.5em}
    \caption{Comparison of \ourmodel against baselines on tasks which include dynamic covariates from the energy and retail domains. For retail, we consider the domain-appropriate WAPE metric.}
    \vspace{-1em}
\end{figure}

\begin{figure}[ht]
    \centering
    \includegraphics[width=0.9\linewidth]{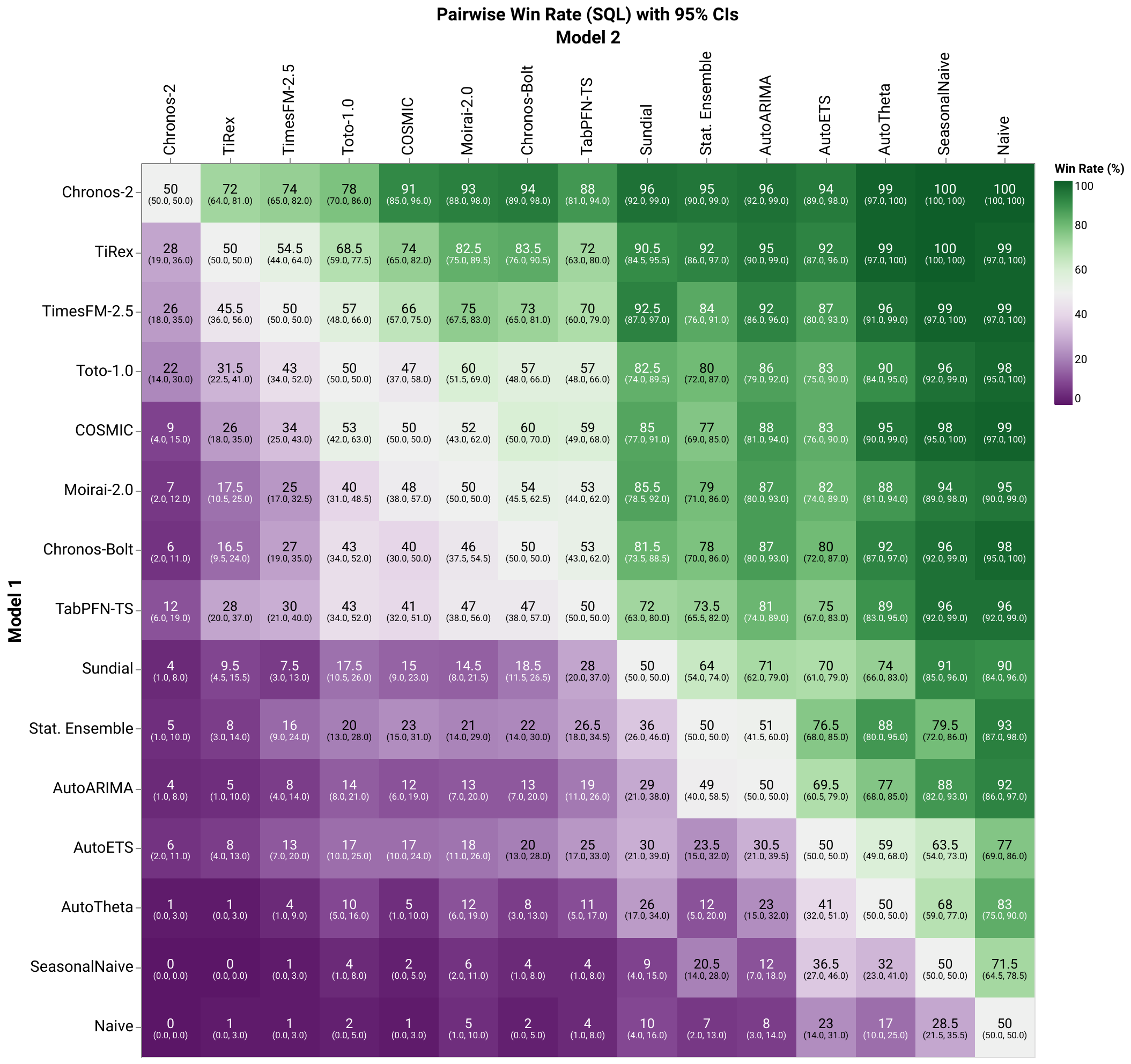}
    \caption{The pairwise win rates for all models on \fevbench with 95\% confidence intervals
(CIs) with respect to SQL metric.}
    \label{fig:fev-pairwise-win-sql}
\end{figure}

\begin{figure}[ht]
    \centering
    \includegraphics[width=0.9\linewidth]{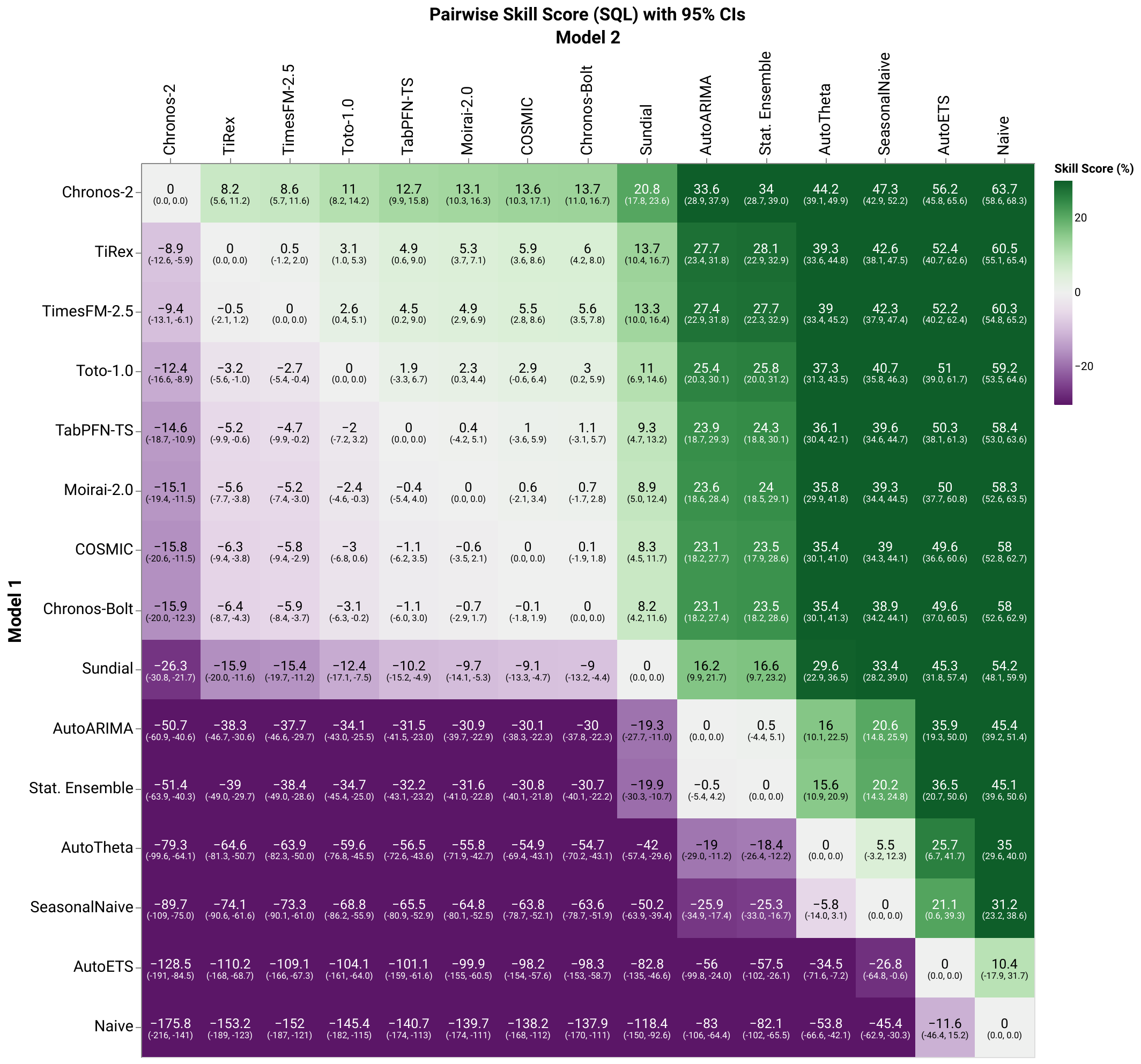}
    \caption{The pairwise skill scores for all models on \fevbench with 95\% confidence intervals
(CIs) with respect to SQL metric.}
    \label{fig:fev-pairwise-skill-sql}
\end{figure}

\begin{figure}[ht]
    \centering
    \includegraphics[width=0.9\linewidth]{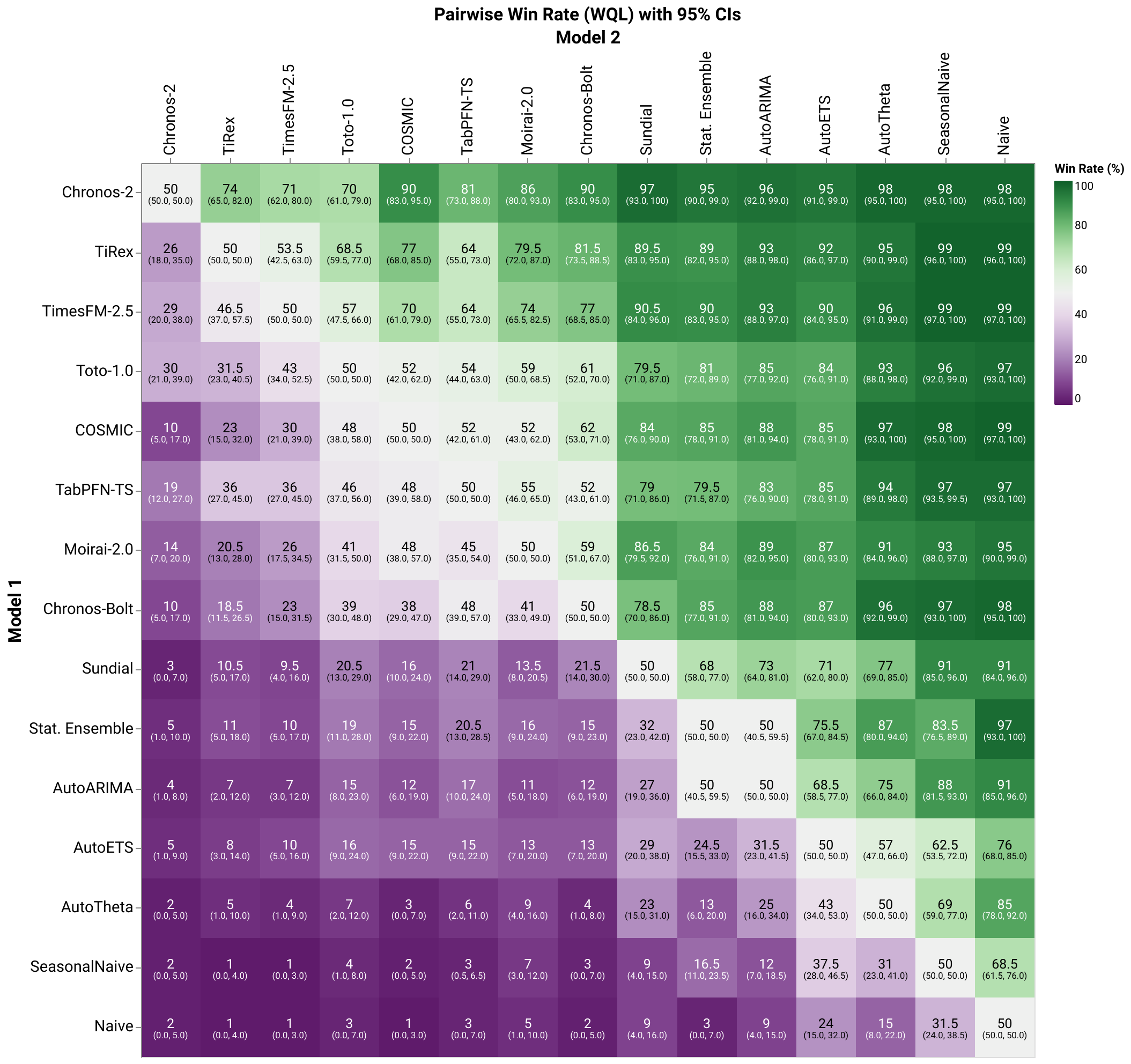}
    \caption{The pairwise win rates for all models on \fevbench with 95\% confidence intervals
(CIs) with respect to WQL metric.}
    \label{fig:fev-pairwise-win-wql}
\end{figure}

\begin{figure}[ht]
    \centering
    \includegraphics[width=0.9\linewidth]{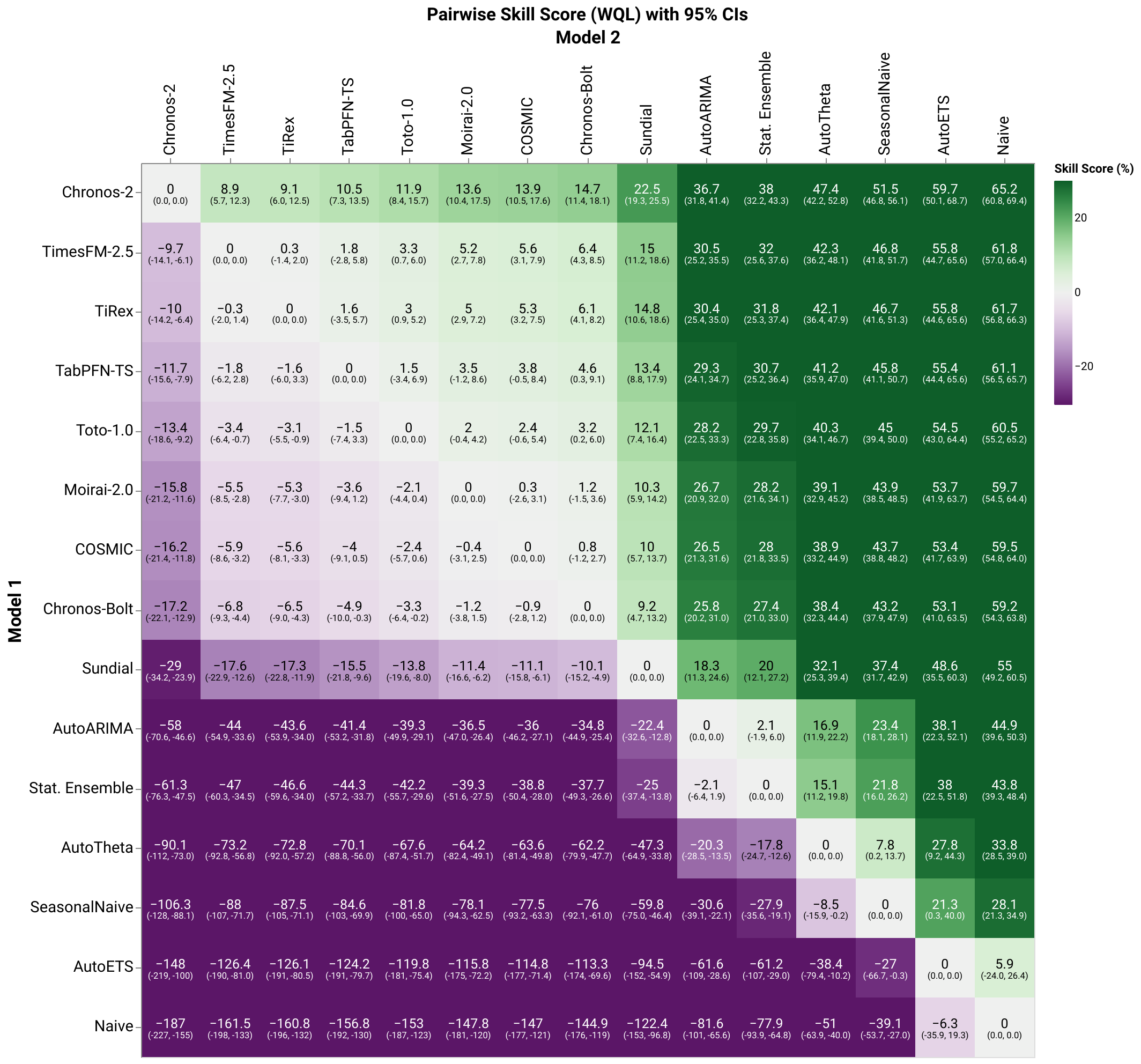}
    \caption{The pairwise skill scores for all models on \fevbench with 95\% confidence intervals
(CIs) with respect to WQL metric.}
    \label{fig:fev-pairwise-skill-wql}
\end{figure}

\begin{figure}[ht]
    \centering
    \includegraphics[width=0.9\linewidth]{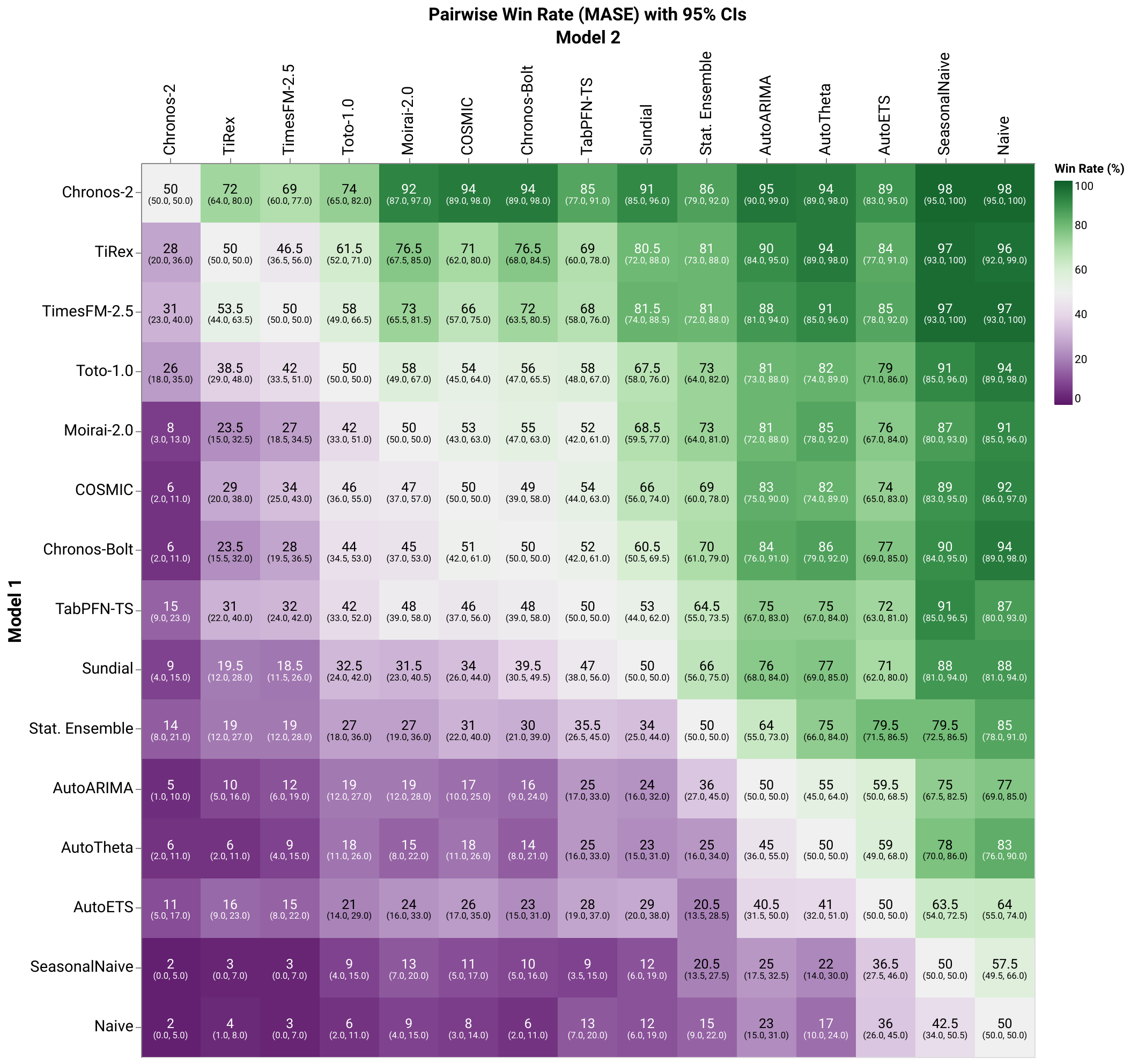}
    \caption{The pairwise win rates for all models on \fevbench with 95\% confidence intervals
(CIs) with respect to MASE metric.}
    \label{fig:fev-pairwise-win-mase}
\end{figure}

\begin{figure}[ht]
    \centering
    \includegraphics[width=0.9\linewidth]{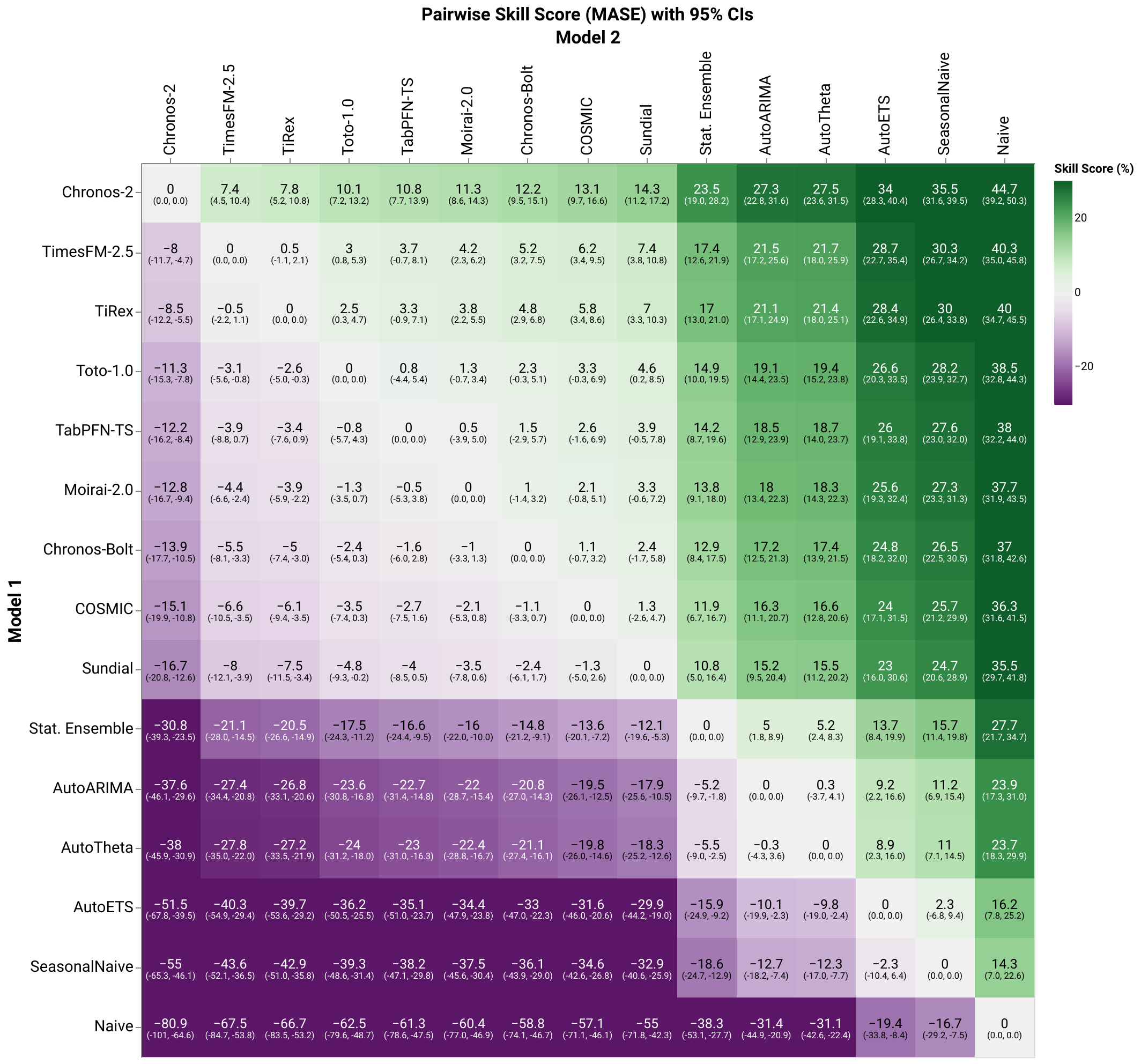}
    \caption{The pairwise skill scores for all models on \fevbench with 95\% confidence intervals
(CIs) with respect to MASE metric.}
    \label{fig:fev-pairwise-skill-mase}
\end{figure}

\begin{figure}[ht]
    \centering
    \includegraphics[width=0.9\linewidth]{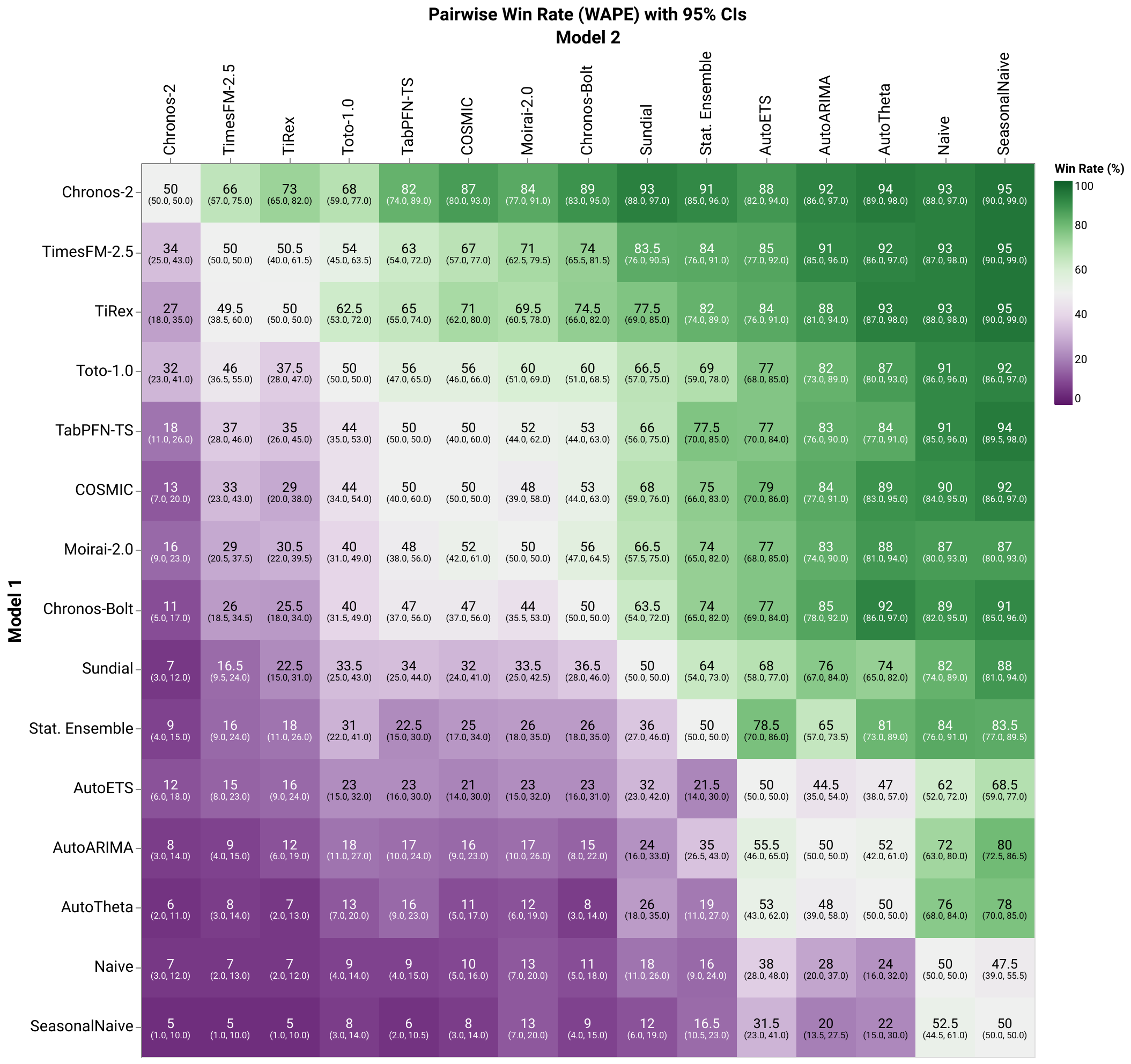}
    \caption{The pairwise win rates for all models on \fevbench with 95\%
(CIs) with respect to WAPE metric.}
    \label{fig:fev-pairwise-win-wape}
\end{figure}

\begin{figure}[ht]
    \centering
    \includegraphics[width=0.9\linewidth]{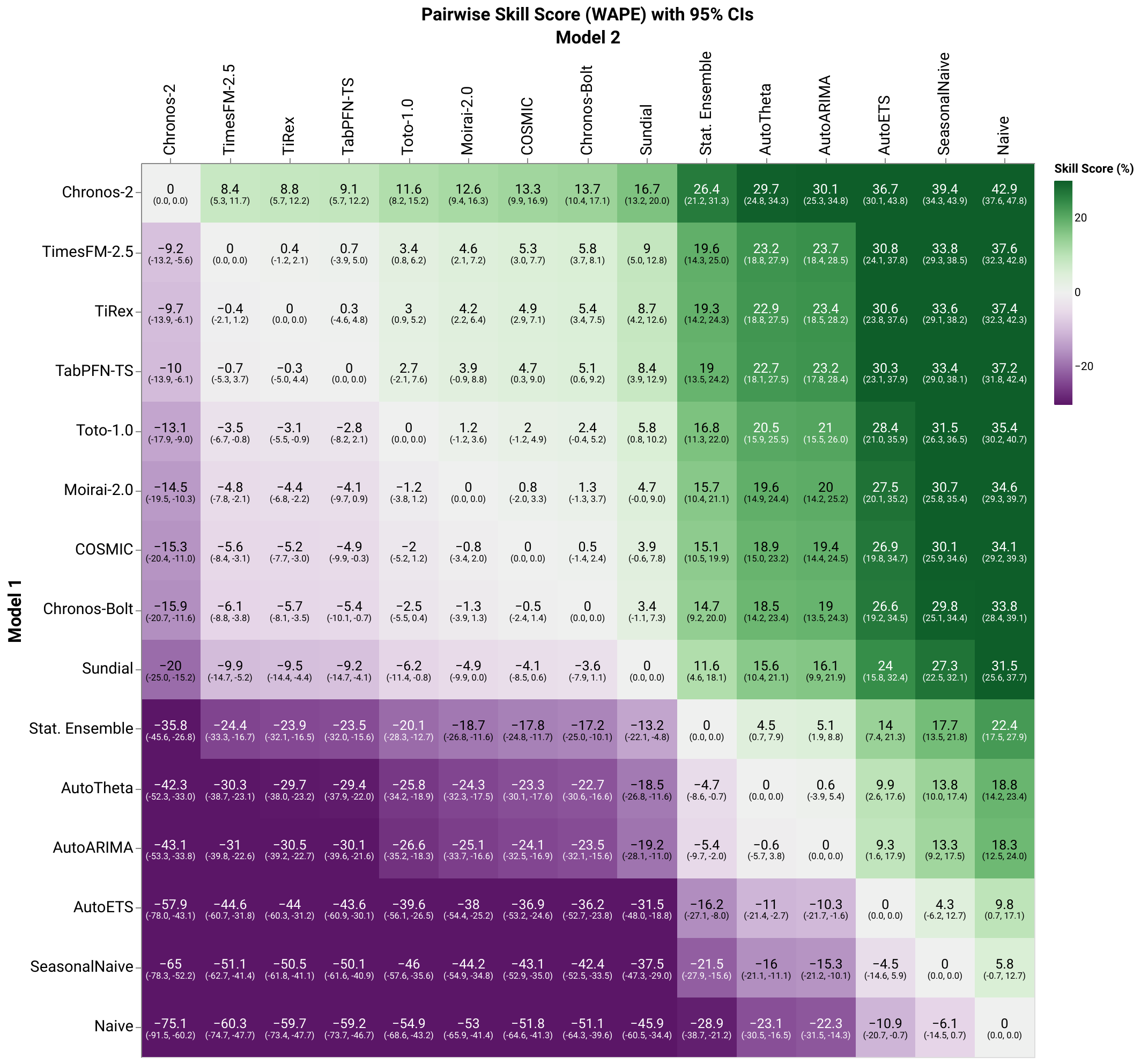}
    \caption{The pairwise skill scores for all models on \fevbench with 95\% confidence intervals
(CIs) with respect to WAPE metric.}
    \label{fig:fev-pairwise-skill-wape}
\end{figure}

\clearpage

\begin{table}[h]
    \resizebox{\textwidth}{!}{%
    \begin{tabular}{llrrrrrrrr}
    \toprule
    \textbf{Task} & \textbf{Freq.} & $H$ & $W$ & \textbf{Median length} & \textbf{\# series} & \textbf{\# targets} & \textbf{\# past cov.} & \textbf{\# known cov.} & \textbf{\# static cov.} \\
    \midrule
    ENTSO-e Load & 15T & 96 & 20 & 175,292 & 6 & 1 & 0 & 3 & 0 \\
    ENTSO-e Load & 30T & 96 & 20 & 87,645 & 6 & 1 & 0 & 3 & 0 \\
    ENTSO-e Load & H & 168 & 20 & 43,822 & 6 & 1 & 0 & 3 & 0 \\
    EPF-BE & H & 24 & 20 & 52,416 & 1 & 1 & 0 & 2 & 0 \\
    EPF-DE & H & 24 & 20 & 52,416 & 1 & 1 & 0 & 2 & 0 \\
    EPF-FR & H & 24 & 20 & 52,416 & 1 & 1 & 0 & 2 & 0 \\
    EPF-NP & H & 24 & 20 & 52,416 & 1 & 1 & 0 & 2 & 0 \\
    EPF-PJM & H & 24 & 20 & 52,416 & 1 & 1 & 0 & 2 & 0 \\
    GFC12 & H & 168 & 10 & 39,414 & 11 & 1 & 0 & 1 & 0 \\
    GFC14 & H & 168 & 20 & 17,520 & 1 & 1 & 0 & 1 & 0 \\
    GFC17 & H & 168 & 20 & 17,544 & 8 & 1 & 0 & 1 & 0 \\
    Solar with Weather & 15T & 96 & 20 & 198,600 & 1 & 1 & 2 & 7 & 0 \\
    Solar with Weather & H & 24 & 20 & 49,648 & 1 & 1 & 2 & 7 & 0 \\
    KDD Cup 2022 & D & 14 & 10 & 243 & 134 & 1 & 9 & 0 & 0 \\
    KDD Cup 2022 & 10T & 288 & 10 & 35,279 & 134 & 1 & 9 & 0 & 0 \\
    KDD Cup 2022 & 30T & 96 & 10 & 11,758 & 134 & 1 & 9 & 0 & 0 \\
    \bottomrule
    \end{tabular}%
    }
    \caption{Subset of datasets from \fevbench with dynamic covariates for the energy domain case study.}
    \label{tab:energy-tasks}
\end{table}

\begin{table}[h]
    \resizebox{\textwidth}{!}{%
    \begin{tabular}{llrrrrrrrr}
    \toprule
    \textbf{Task} & \textbf{Freq.} & $H$ & $W$ & \textbf{Median length} & \textbf{\# series} & \textbf{\# targets} & \textbf{\# past cov.} & \textbf{\# known cov.} & \textbf{\# static cov.} \\
    \midrule
    Favorita Store Sales & M & 12 & 2 & 54 & 1,579 & 1 & 1 & 1 & 6 \\
    Favorita Store Sales & W & 13 & 10 & 240 & 1,579 & 1 & 1 & 1 & 6 \\
    Favorita Store Sales & D & 28 & 10 & 1,688 & 1,579 & 1 & 1 & 2 & 6 \\
    Favorita Transactions & M & 12 & 2 & 54 & 51 & 1 & 1 & 0 & 5 \\
    Favorita Transactions & W & 13 & 10 & 240 & 51 & 1 & 1 & 0 & 5 \\
    Favorita Transactions & D & 28 & 10 & 1,688 & 51 & 1 & 1 & 1 & 5 \\
    M5 & M & 12 & 1 & 58 & 30,490 & 1 & 0 & 8 & 5 \\
    M5 & W & 13 & 1 & 257 & 30,490 & 1 & 0 & 8 & 5 \\
    M5 & D & 28 & 1 & 1,810 & 30,490 & 1 & 0 & 8 & 5 \\
    Rohlik Orders & W & 8 & 5 & 170 & 7 & 1 & 9 & 4 & 0 \\
    Rohlik Orders & D & 61 & 5 & 1,197 & 7 & 1 & 9 & 4 & 0 \\
    Rohlik Sales & W & 8 & 1 & 150 & 5,243 & 1 & 1 & 13 & 7 \\
    Rohlik Sales & D & 14 & 1 & 1,046 & 5,390 & 1 & 1 & 13 & 7 \\
    Rossmann & W & 13 & 8 & 133 & 1,115 & 1 & 1 & 4 & 10 \\
    Rossmann & D & 48 & 10 & 942 & 1,115 & 1 & 1 & 5 & 10 \\
    Walmart & W & 39 & 1 & 143 & 2,936 & 1 & 0 & 10 & 4 \\
    Hermes & W & 52 & 1 & 261 & 10,000 & 1 & 0 & 1 & 2 \\
    \bottomrule
    \end{tabular}%
    }
    \caption{Subset of datasets from \fevbench with dynamic covariates for the retail domain case study.}
    \label{tab:retail-tasks}
\end{table}

\end{document}